\documentclass[journal]{IEEEtran}

\usepackage[utf8]{inputenc}
\usepackage{color}
\usepackage{xcolor}
\usepackage{array}
\usepackage{verbatim}
\usepackage{float}
\usepackage{amsmath}
\usepackage{amsthm}
\usepackage{amssymb}
\usepackage{graphicx}
\usepackage{longtable}
\usepackage{multirow}
\usepackage{booktabs}
\usepackage[unicode=true,
bookmarks=false,
breaklinks=false,pdfborder={0 0 1},colorlinks=false]
{hyperref}
\hypersetup{
	colorlinks,bookmarksopen,bookmarksnumbered,citecolor=blue,urlcolor=blue}
\usepackage{cite}

\usepackage{lipsum}
\usepackage{mathtools}
\usepackage{cuted}
\providecommand{\tabularnewline}{\\}
\usepackage{algorithmic}
\usepackage{longtable}

\floatstyle{ruled}
\newfloat{algorithm}{tbp}{loa}
\providecommand{\algorithmname}{Algorithm}
\floatname{algorithm}{\protect\algorithmname}

\makeatletter
\let\oldforeign@language\foreign@language
\DeclareRobustCommand{\foreign@language}[1]{%
	\lowercase{\oldforeign@language{#1}}}

\let\oldforeign@language\foreign@language
\DeclareRobustCommand{\foreign@language}[1]{%
	\lowercase{\oldforeign@language{#1}}}

\ifCLASSINFOpdf
\else
\fi

\@ifundefined{showcaptionsetup}{}{%
	\PassOptionsToPackage{caption=false}{subfig}}
\usepackage{subfig}

\usepackage{balance}

\ifCLASSINFOpdf
\else
\fi

\newtheorem{lem}{Lemma}

\newtheorem{thm}{Theorem}

\newtheorem{assum}{Assumption}
\ifCLASSINFOpdf
\else
\fi

\def\ps@IEEEtitlepagestyle{%
	\def\@oddhead{\parbox[t][\height][t]{\textwidth}{\centering \scriptsize
			Personal use of this material is permitted. Permission from the author(s) and/or copyright holder(s), must be obtained for all other uses. Please contact us and provide details if you believe this document breaches copyrights.\\
			\noindent\makebox[\linewidth]{}
		}\hfil\hbox{}}%
	\def\@evenhead{\scriptsize\thepage \hfil \leftmark\mbox{}}%
	\def\@oddfoot{\parbox[t][\height][l]{\textwidth}{
			\vspace{-20pt}{\rule{\textwidth}{0.4pt}}\\ \footnotesize			{\bf{\footnotesize\textcolor{red}{A. Shevidi and H. A. Hashim, "Quaternion-based Adaptive Backstepping Fast Terminal Sliding Mode Control for Quadrotor UAVs with Finite Time Convergence," Results in Engineering, pp. 102497, 2024.}}} doi: \href{https://doi.org/10.1016/j.rineng.2024.102497}{10.1016/j.rineng.2024.102497}\\
			\noindent\makebox[\linewidth]
		}\hfil\hbox{}}%
	\def\@evenfoot{\MYfooter}}

\makeatother
\begin{document}
	\bstctlcite{IEEEexample:BSTcontrol}

\title{Quaternion-based Adaptive Backstepping Fast Terminal Sliding Mode Control for Quadrotor UAVs with Finite Time Convergence}

\author{Arezo Shevidi and Hashim A. Hashim
	\thanks{This work was supported in part by National Sciences and Engineering
		Research Council of Canada (NSERC), under the grants RGPIN-2022-04937
		and DGECR-2022-00103.}
	\thanks{A. Shevidi and H. A. Hashim are with the Department of Mechanical
		and Aerospace Engineering, Carleton University, Ottawa, Ontario, K1S-5B6,
		Canada, (e-mail: hhashim@carleton.ca).}
}



\maketitle
\begin{abstract}
This paper proposes a novel quaternion-based approach for tracking
the translation (position and linear velocity) and rotation (attitude
and angular velocity) trajectories of underactuated Unmanned Aerial
Vehicles (UAVs). Quadrotor UAVs are challenging regarding accuracy,
singularity, and uncertainties issues. Controllers designed based
on unit-quaternion are singularity-free for attitude representation
compared to other methods (e.g., Euler angles), which fail to represent
the vehicle's attitude at multiple orientations. Quaternion-based
Adaptive Backstepping Control (ABC) and Adaptive Fast Terminal Sliding
Mode Control (AFTSMC) are proposed to address a set of challenging
problems. A quaternion-based ABC, a superior recursive approach, is
proposed to generate the necessary thrust handling unknown uncertainties
and UAV translation trajectory tracking. Next, a quaternion-based
AFTSMC is developed to overcome parametric uncertainties, avoid singularity,
and ensure fast convergence in a finite time. Moreover, the proposed
AFTSMC is able to significantly minimize control signal chattering,
which is the main reason for actuator failure and provide smooth and
accurate rotational control input. To ensure the robustness of the
proposed approach, the designed control algorithms have been validated
considering unknown time-variant parametric uncertainties and significant
initialization errors. The proposed techniques has been compared to
state-of-the-art control technique. 
\end{abstract}


\section{Introduction}\label{sec1}

\subsection{Motivation}
\IEEEPARstart{I}{n} recent years, Unmanned Aerial Vehicles (UAVs) application, commonly
known as drones, has witnessed a surge in research interest. These
cutting-edge advanced devices have captured significant attention
due to remarkable adaptability, low energy consumption, and small
size \cite{hashim2021geometricSLAM,chen2014planning,cheng2023ai,hashim2023uwbIEEEITS,khan2015information,bardera2023numerical}.
The convergence of breakthroughs in electric power storage, wireless
communication \cite{huo2019distributed}, agriculture, space exploration,
and firefighting has been pivotal in propelling the development of
innovative drone models \cite{huo2019distributed,hashim2023uwbIEEEITS,cheng2023ai,bartolomei2023fast,jonnalagadda2024segnet}.
Among various kinds of drones, quadrotors represent a nascent and
highly versatile category of UAVs, showing interesting advancements
in diverse applications. Quadrotors are used in various fields, from
logistics, delivery services, and environmental conservation efforts
to search and rescue operations, building inspection and reconnaissance,
maintenance tasks, precision agriculture, videography and photography,
mining operations, as well as hazardous activities \cite{hassanalian2017classifications,hashim2023uwbIEEEITS,jonnalagadda2024segnet,hashim2023observer,cheng2023ai}.
Moreover, quadrotors provide remarkable adaptability for indoor and
outdoor flights, enabling deployment in underwater environments, over
water surfaces, terrestrial landscapes, airspace, and even outer space,
further enhancing its operational flexibility \cite{hassanalian2017classifications}.
Quadrotors hold inherent positive features that distinguish them from
conventional helicopters, making them in high demand for various practical
applications \cite{hashim2023exponentially,hashim2023observer}. The
features stem from the lightweight structure and streamlined mechanical
system, enabling them to guarantee stable hover flight, which is a
crucial prerequisite in real applications. Such attributes not only
optimize flight performance but also improve operational safety \cite{hashim2023exponentially,wanner2024uav,hashim2023observer}.
Moreover, quadrotors  can maneuver precisely in close quarters, enabling
them to perform vertical take-offs and navigate through cluttered
and confined spaces \cite{du2023fault}. As a result, quadrotors serve
as invaluable aids in various daily tasks and prove particularly indispensable
in tackling complex and hazardous missions \cite{du2023fault}. However,
despite their many advantages, quadrotors encounter significant challenges
primarily linked to the rotation of their propellers and the flapping
of their blades \cite{wang2016trajectory}. Furthermore, the strong
interdependence between their position and attitude amplifies the
complexity of their control systems, characterized by under-actuation
and susceptibility to disturbances arising from aerodynamic effects,
including sensitivity to wind gusts, thereby compromising stability
\cite{liu2017robust,du2023fault}.



\subsection{Related Work and Challenges}

The control of UAVs presents notable limitations and challenges that
arise from model representations and control design. Several control
strategies and approaches have been investigated to tackle control
system limitations, including singularity, stability, and chattering
\cite{ryan2013lmi,lopez2023pid,hashim2023observer,martins2019linear,mofid2022desired,lindqvist2020nonlinear}.
For instance,\cite{lopez2023pid} has proposed a conventional PID-based
controller for attitude control. Also, in \cite{willis2020state},
state-dependent Linear Quadratic Regulator (LQR) control is suggested.
In\cite{lindqvist2020nonlinear}, nonlinear model predictive control
(NMPC) is developed to navigate and avoid obstacles for UAVs. In \cite{hashim2023observer,hashim2023exponentially},
nonlinear cascaded vision-based controllers that relies on Lyapunov
stability have been developed. In \cite{saccani2022multitrajectory},
multi-trajectory MPC is explored for UAV navigation while the environment
is unknown. Authors investigate robust MPC to handle the mismatch
uncertainty of the model or external disturbance. They designed the
robust MPC by defining optimization problems with constraints in \cite{zhang2021robust,saccani2022multitrajectory,ali2024mpc}.
Although the MPC approach is an advanced control technique based on
prediction and optimization, its implementation is not cost-effective
in practical application. Moreover, the computational burden is the
major disadvantage of NMPC. In addition, the above-listed controllers
are not characterized with finite-time convergence. Among the mentioned
control strategies, Sliding Mode Control (SMC) has gained significant
attention regarding simplicity, accuracy, and handling uncertainties.
In \cite{vahdanipour2019adaptive,baek2016new}, SMC is utilized to
control UAV systems, robotics, navigation, and aircraft guidance.
For example, in \cite{vahdanipour2019adaptive}, the work developed
a fractional order SMC for a quadrotor with varying load. The main
point of the controller is to provide missile guidance in the presence
of uncertainty \cite{vahdanipour2019adaptive}. In \cite{baek2016new},
the authors use adaptive SMC to control robot manipulators.

SMC, despite its streamlined implementation and robustness, holds
critical drawbacks such as (1) Chattering (2) Finite-time issues (3)
Performance during reaching phase \cite{derakhshannia2022disturbance,pisano2011sliding,mobayen2014design,munoz2019adaptive}.
Chattering and oscillation caused by switching control action can
be a critical issue in SMC. In practical flight applications, chattering
can be considered one of the negative consequences of SMC because
of its critical impact on the stability and robustness of an actuator
system \cite{derakhshannia2022disturbance,said2023application}. Regarding
the finite time issue, SMC may not guarantee convergence to desired
states in the finite time since the sliding surface is designed in
a linear form. In other words, while the system gradually reaches
the sliding surface, the convergence time may even be unbounded depending
on the specific initial conditions. Additionally, according to\cite{pisano2011sliding,mobayen2014design},
during the reaching phase, SMC may not be robust, and the system may
be susceptible to instability in the presence of external disturbances.
Moreover, convergence can be more complicated if the states are not
close to equilibrium points \cite{mobayen2014design}. To improve
the performance of controllers, advanced sliding mode techniques have
been developed to remove chattering, ensure finite time convergence,
and enhance performance in \cite{labbadi2019robust,labbadi2020path,mofid2021adaptive,mofid2022desired,zheng2023adaptive,sanwale2023robust,ahmadi2023active,gao2022adaptive,xie2023fixed}.
In \cite{labbadi2019robust}, a fast terminal sliding mode control
is designed to reduce the effect of chattering while the quadrotor
is under uncertainties. The work in \cite{labbadi2020path} implemented
super twisting SMC to improve the performance of the fully-actuated
UAVs in speed tracking with fast convergence. In \cite{zheng2023adaptive},
more efforts have been made to enhance stability, while uncertainty
and external disturbances exist in UAVs. The fractional order non-singular
TSMC with adaptation laws is used to overcome external disturbances
and uncertainties for advanced layout carrier-based UAV \cite{zheng2023adaptive}.
\cite{sanwale2023robust} has explored dynamic SMC with adaptive control
which is tolerable in the presence of faults and fixed-time disturbance.
The work in \cite{ahmadi2023active} investigated the mixture of the
nonlinear observer with SMC to overcome UAV system faults. The work
in \cite{gao2022adaptive,xie2023fixed} combines neural networks with
FTSMC to design fault-tolerant control for UAVs.

Tackling underactuation represents the first challenge. The majority
of the UAVs are considered to be underactuated systems, which is a
fundamental challenge for achieving robust and reliable missions \cite{hashim2023exponentially}.
From realistic and practical points of view, having control over all
states is often impossible. Due to such a fundamental deficiency,
stabilizing and achieving desired tasks become a complex challenge.
Unfortunately, most of the above-listed control systems have been
designed for fully-actuated UAVs, which is a fundamental challenge
(e.g., an adaptive and super-twisting sliding mode controller \cite{mofid2022desired}).
However, such an assumption is not realistic nor practical for the
majority of UAVs. In this regard, backstepping is an effective approach
to address under-actuation issues. The main idea of backstepping is
to generate an intermediary control input (virtual control), done
in a cascaded way to tackle the limitation of underactuated control
system \cite{kwan2000robust}. The second challenge is avoiding singularity
in each control design and attitude representation to ensure UAV stabilization.
Although the above-listed techniques have attempted to avoid singularity
and chattering, the designed controllers are based on Euler angles,
which can lead to a significant issue in tracking and UAV stability
\cite{hashim2023observer}. Unit-quaternion can be an alternative
method to represent attitude without singularity issues. In particular,
Euler angles hold significant shortcomings, such as kinematic singularities
and local representation \cite{hashim2019special}. Such an issue
causes the controller to fail, resulting in undesired performance.

\subsection{Contributions}


In this paper, we consider cascaded control design and adopt quaternion-based
ABC to handle the UAV underactuated issues leveraging an auxiliary
control input (virtual control). The auxiliary control of ABC generates
the required thrust and the desired UAV orientation for attitude control.
In addition, with the novel quaternion-based controllers, we resolve
the persistent singularity shortcomings of other common methods in
literature (e.g. Euler-based controller). Afterwards, we mitigate
chattering issues of the control signal, the fundamental reason for
actuator damage, by modifying the sliding surface. The ``fast''
and adaptive aspects of the proposed quaternion-based AFTSMC are crucial
in reducing chattering problems. Also, The finite-time convergences
with smooth control signals are guaranteed by the established quaternion-based
AFTSMC compared to conventional Euler-based SMC. In order to enhance
the robustness of the proposed controllers, the adaptive features
of proposed controllers are developed to adjust control parameters
while unknown time-varying parametric uncertainties and significant
initialization errors exist. The main contributions are as follows: 
\begin{itemize}
	\item[(1)] A novel quaternion-based ABC is proposed to control the UAV translation
	dynamics by addressing the underactuated issues via an auxiliary control
	input (virtual control). 
	\item[(2)] A quaternion-based AFTSMC is proposed for the UAV rotational dynamics
	to mitigate the chattering (the main reason for actuator failure)
	by fast and non-singular features of the sliding surface with adaptation
	control parameters. 
	\item[(3)] The proposed attitude quaternion-based controller guarantees states
	converge to desired values in finite time with smooth control signal. 
	\item[(4)] The innovative quaternion-based controllers addresses the persisting
	challenge of Euler angles-based solutions, namely kinematic singularity
	and failure of model representation at certain configuration. 
\end{itemize}
The novel control systems utilize adaptation mechanisms and guarantee
asymptotic stability using the Barbalet Lemma and Lyapunov theorem.
The simplicity and straightforward design approach is not only safer
but also more autonomous in flight missions compared to other methods
in the literature.

A comparison to state-of-the-art controllers for UAVs has been included.

\subsection{Structure}

\begin{table}[!t]
	\centering{}\caption{\label{tab:Table-of-Notations2}Nomenclature}
	\begin{tabular}{ll>{\raggedright}p{5.1cm}}
		\toprule 
		\addlinespace
		$\{\mathcal{B}\}$  & :  & Body-frame (moving-frame)\tabularnewline
		\addlinespace
		$\{\mathcal{I}\}$  & :  & Inertial-frame (fixed-frame)\tabularnewline
		\addlinespace
		$I_{xx}$, $I_{yy}$, $I_{zz}$  & :  & UAV inertia \tabularnewline
		\addlinespace
		$J$  & :  & Inertia matrix \tabularnewline
		\addlinespace
		$m$  & :  & Mass of the UAV \tabularnewline
		\addlinespace
		$g$  & :  & Gravitational acceleration \tabularnewline
		\addlinespace
		$X=[x_{p},y_{p},z_{p}]^{\top}$  & :  & True position of the UAV \tabularnewline
		\addlinespace
		$X_{d}=[x_{d},y_{d},z_{d}]^{\top}$  & :  & Desired position of the UAV \tabularnewline
		\addlinespace
		$V=[v_{1},v_{2},v_{3}]^{\top}$  & :  & Linear velocity of the UAV \tabularnewline
		\addlinespace
		$R$  & :  & Rotational matrix (True UAV orientation)\tabularnewline
		\addlinespace
		$\Omega_{B}=[p_{v},q_{v},r_{v}]^{\top}$  & :  & UAV angular velocity in body-frame\tabularnewline
		\addlinespace
		$\Omega_{d}$  & :  & Desired angular velocity \tabularnewline
		\addlinespace
		$\Omega_{e}$  & :  & Angular velocity error \tabularnewline
		\addlinespace
		$R_{e}$  & :  & Rotational matrix error \tabularnewline
		\addlinespace
		$Q$  & :  & Unit quaternion (True UAV orientation)\tabularnewline
		\addlinespace
		$Q_{d}$  & :  & Desired Unit quaternion \tabularnewline
		\addlinespace
		$Q_{e}=[e_{0},e^{\top}]$  & :  & Quaternion error \tabularnewline
		\addlinespace
		$\mathcal{T}=[\mathcal{T}_{1},\mathcal{T}_{2},\mathcal{T}_{3}]$  & :  & Rotational torque input\tabularnewline
		\addlinespace
		$F_{th}$  & :  & Total thrust \tabularnewline
		\addlinespace
		$e_{xp}$, $e_{yp}$, $e_{zp}$  & :  & Position error\tabularnewline
		\addlinespace
		$e_{v1}$, $e_{v2}$, $e_{v3}$  & :  & Linear velocity error\tabularnewline
		\addlinespace
		$\hat{n}_{x2}$, $\hat{n}_{y2}$, $\hat{n}_{z2}$  & :  & Adaptive estimate (position control parameters) \tabularnewline
		\addlinespace
		$\hat{k}_{1}$, $\hat{k}_{2}$, $\hat{k}_{3}$  & :  & Adaptive estimate (attitude control parameters) \tabularnewline
		\addlinespace
		$s$  & :  & FTSMC surface \tabularnewline
		\addlinespace
		$U_{x}$, $U_{y}$, $U_{z}$  & :  & Virtual control\tabularnewline
		\bottomrule
	\end{tabular}
\end{table}

The remaining of the paper is summarized as follows: In Section \ref{sec:pre},
the preliminaries and math notation are provided. Section \ref{sec:dyna}
presents the UAV model in quaternion representation. Section \ref{sec:Control}
introduces problem formulation and proposes quaternion-based controllers.
Section \ref{sec:sum} introduces the implementation steps. Section
\ref{sec:Result} depicts the simulation results confirming the effectiveness
of the proposed controller. Finally, section \ref{sec:co} summarizes
the paper.

Some of the commonly used notation is provided in Table \ref{tab:Table-of-Notations2}.

\section{Preliminaries\label{sec:pre}}

\subsection{Math Notation}

In this paper, $\mathbb{R}$ describes the set of real numbers. $\mathbb{R}^{c\times d}$
represents the set of real-numbers with dimensional space $c$-by-$d$.
$\boldsymbol{I}_{c}\in\mathbb{R}^{c\times c}$ and $\boldsymbol{0}_{c}\in\mathbb{R}^{c\times c}$
represent identity and zero matrices, respectively. $\|y\|=\sqrt{y^{\top}y}$
refers to an Euclidean norm of the column vector $y\in\mathbb{R}^{n}$.
Let $[v]_{\times}$ denote a skew-symmetric matrix such that 
\begin{align*}
	\left[v\right]_{\times} & =\left[\begin{array}{ccc}
		0 & -v_{3} & v_{2}\\
		v_{3} & 0 & -v_{1}\\
		-v_{2} & v_{1} & 0
	\end{array}\right],\hspace{1em}v=\left[\begin{array}{c}
		v_{1}\\
		v_{2}\\
		v_{3}
	\end{array}\right]
\end{align*}
For $M\in\mathbb{R}^{3\times3}$ and $w,v\in\mathbb{R}^{3}$, the
following characteristics hold true \cite{hashim2023observer,hashim2023exponentially}:
\begin{equation}
	\begin{aligned}-[v]_{\times}v & =0_{3\times1}\\{}
		[w]_{\times}^{\top} & =-[w]_{\times}\\{}
		[w]_{\times}v & =-v[w]_{\times}\\{}
		[Mv]_{\times} & =M[v]_{\times}M^{\top}
	\end{aligned}
	\label{eq:eqOne}
\end{equation}
In this work, we consider a UAV travelling in 3D space where the vehicle's
moving-frame (body-frame) is denoted by $\{\mathcal{B}\}=\{x_{B},y_{B},z_{B}\}$
while the inertial-frame is described by $\{\mathcal{I}\}=\{x_{p},y_{p},z_{p}\}$.

\subsection{Unit-quaternion}

Unit-quaternion is a useful tool for singularity-free attitude (orientation)
representation in 3D space. $Q$ is a unit-quaternion vector such
that $Q=[q_{0},q_{1},q_{2},q_{3}]^{\top}=[q_{0},\boldsymbol{q}^{\top}]^{\top}\in\mathbb{R}^{4}$
and $||Q||=1$ \cite{hashim2019special,hashim2023exponentially}  where
$q_{0}\in\mathbb{R}$ and $\boldsymbol{q}=[q_{1},q_{2},q_{3}]^{\top}\in\mathbb{R}^{3}$.
The unit-quaternion four elements are given as follows \cite{hashim2019special}:
\[
Q=q_{0}+iq_{1}+jq_{2}+kq_{3}
\]
where $i$, $j$, and $k$ refers to the standard basis-vectors. $Q^{-1}$
denoted inversion of $Q$ and is given by \cite{hashim2023observer}:
\[
Q^{-1}=Q^{*}=[q_{0},-q_{1},-q_{2},-q_{3}]^{\top}=[q_{0},-\boldsymbol{q}^{\top}]^{\top}\in\mathbb{R}^{4}
\]
Consider $\otimes$ to be quaternion multiplication and let $Q=[q_{0},\boldsymbol{q}^{\top}]^{\top}$
and $Q_{d}=[q_{0d},\boldsymbol{q}_{d}^{\top}]=[q_{0d},q_{1d},q_{2d},q_{3d}]^{\top}$
be two quaternion vectors describing true and desired UAV attitude,
respectively, where $q_{0},q_{0d}\in\mathbb{R}$ and $\boldsymbol{q},\boldsymbol{q}_{d}\in\mathbb{R}^{3}$.
The quaternion multiplication of $Q$ and $Q_{d}$ is given by: 
\[
Q\otimes Q_{d}=\left[\begin{array}{c}
	q_{0}q_{0d}+\boldsymbol{q}^{\top}\boldsymbol{q}_{d}\\
	q_{0d}\boldsymbol{q}-q_{0}\boldsymbol{q}_{d}+[\boldsymbol{q}]_{\times}\boldsymbol{q}_{d}
\end{array}\right]=\left[\begin{array}{c}
	e_{0}\\
	e
\end{array}\right]
\]
Quaternion multiplication is associative but not commutative. The
UAV attitude representation in form of a rotational matrix described
with respect to the Lie Group of the Special Orthogonal Group $SO(3)$
is described by 
\[
R=(q_{0}^{2}-\boldsymbol{q}^{\top}\boldsymbol{q})\mathbf{I}_{3}+2\boldsymbol{q}\boldsymbol{q}^{\top}+2q_{0}[\boldsymbol{q}]_{\times}\in SO(3)\subset\mathbb{R}^{3\times3}
\]
Which is equivalent 
\begin{equation}
	R=\left[\begin{array}{lll}
		1-2\left(q_{2}^{2}+q_{3}^{2}\right) & 2\left(q_{1}q_{2}-q_{0}q_{3}\right) & 2\left(q_{1}q_{3}+q_{0}q_{2}\right)\\
		2\left(q_{2}q_{1}+q_{0}q_{3}\right) & 1-2\left(q_{1}^{2}+q_{3}^{2}\right) & 2\left(q_{2}q_{3}-q_{0}q_{1}\right)\\
		2\left(q_{3}q_{1}-q_{0}q_{2}\right) & 2\left(q_{3}q_{2}+q_{0}q_{1}\right) & 1-2\left(q_{1}^{2}+q_{2}^{2}\right)
	\end{array}\right]\in SO(3)\label{eq:Rmatrix}
\end{equation}
where $det(R)=+1$ and $RR^{\top}=\mathbf{I}_{3}$ with $det(\cdot)$
being determinant of a matrix and $R\in\{\mathcal{B}\}$.

\section{UAV Dynamical Model and Quaternion Representation\label{sec:dyna}}

\begin{figure}
	\centering{}\includegraphics[scale=0.5]{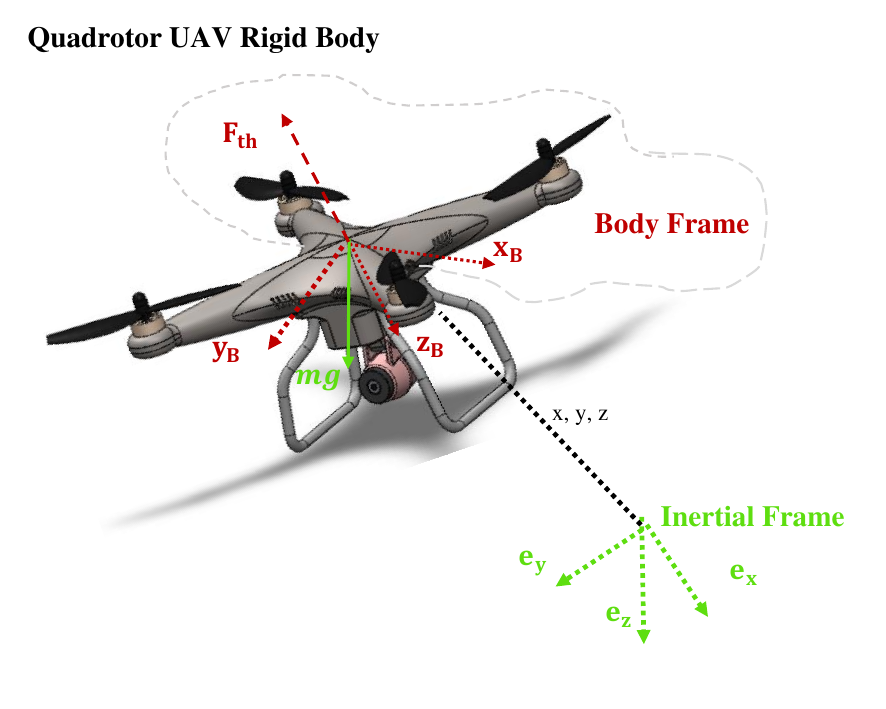}\caption{Quadrotor UAV configuration}
	\label{fig:fig6} 
\end{figure}

The structure of the quadrotor consists of four rotors where the blades
rotate via rotors in such a direction to produce thrust. In this regard,
the flight motion of a UAV consists of two parts, namely, translation
and rotation \cite{labbadi2019robust,hashim2023observer}. The structure
of a quadrotor UAV is presented in Figure \ref{fig:fig6}. Note that
the under-actuation nature of UAV requires a cascaded design of the
controller, translational controller (outer part) and rotational controller
(inner part). In the translation part, the UAV controller is tasked
with thrust generation to follow predefined desired position trajectories
$x_{d}$, $y_{d}$, and $z_{d}$. In the rotation part, the UAV controller
is tasked with generating rotational torques to follow predefined
attitude trajectories $Q_{d}=[q_{0d},q_{1d},q_{2d},q_{3d}]$ produced
by the translational controller (outer part).

\subsection{UAV Translation and Rotational Dynamics}

The UAV translation dynamics are given as follows \cite{hashim2023observer,hashim2023exponentially}
\begin{equation}
	\text{ Translation : }\begin{array}{ll}
		\left[\begin{array}{l}
			\dot{X}\\
			\dot{V}
		\end{array}\right] & =\left[\begin{array}{c}
			V\\
			\left[\begin{array}{l}
				0\\
				0\\
				g
			\end{array}\right]+m^{-1}RU
		\end{array}\right]\end{array}\label{eq:eq1}
\end{equation}
where $m$ and $g$ are mass and gravity acceleration, respectively,
$U=[0,0,-F_{th}]^{\top}\in\mathbb{R}^{3}$ is the total input control
for translation state space equation of UAV and $F_{th}$ is the total
thrust. $X=[x_{p},y_{p},z_{p}]^{\top}\in\mathbb{R}^{3}$ denotes UAV's
true position, $V=[v_{1},v_{2},v_{3}]^{\top}\in\mathbb{R}^{3}$ denotes
UAV's true velocity, and $R\in SO(3)\subset\mathbb{R}^{3\times3}$
denotes UAV true attitude described in \eqref{eq:Rmatrix} where $X,V\in\{\mathcal{I}\}$
and $R\in\{\mathcal{B}\}$. The UAV rotation dynamics are defined
by \cite{hashim2023observer,fu2023iterative,hashim2023exponentially}:
\begin{equation}
	\text{ Rotation}\left\{ \begin{array}{ll}
		\dot{R} & =-[\Omega_{B}]_{\times}R\\
		J\dot{\Omega}_{B} & =[J\Omega_{B}]_{\times}\Omega_{B}+\mathcal{T},
	\end{array}\right.\label{eq:eq4}
\end{equation}
where $\Omega_{B}=[p_{v},q_{v},r_{v}]^{\top}\in\mathbb{R}^{3}$, $R\in SO(3)$,
$\mathcal{T}=[\mathcal{T}_{1},\mathcal{T}_{2},\mathcal{T}_{3}]^{\top}\in\mathbb{R}^{3}$
are UAV's angular velocity, attitude, and torque input, respectively,
with $R,\Omega_{B},\mathcal{T}\in\{\mathcal{B}\}$, and $J\in\mathbb{R}^{3\times3}$
refers to the inertia matrix which is symmetric. Based on \eqref{eq:eq4},
the equivalent attitude dynamics in quaternion form are given by \cite{hashim2023observer,hashim2023exponentially}:
\begin{align}
	\text{Rotation} & \begin{cases}
		\dot{q}_{0} & =-\frac{1}{2}\boldsymbol{q}^{\top}\Omega_{B}\\
		\dot{\boldsymbol{q}} & =\frac{1}{2}\left(q_{0}\Omega_{B}+[\boldsymbol{q}]_{\times}\Omega_{B}\right)\\
		J\dot{\Omega}_{B} & =[J\Omega_{B}]_{\times}\Omega_{B}+\mathcal{T}
	\end{cases}\label{eq:eq6}
\end{align}
with $Q\in\mathbb{S}^{3}$ being UAV's orientation with respect to
unit-quaternion.

\begin{assum}\label{Assum:Assume1} It is assumed that the structure
	of UAV is symmetric and rigid. \end{assum}

\begin{assum}\label{Assum:Assume2} The desired position $X_{d}=[x_{d},y_{d},z_{d}]^{\top}\in\mathbb{R}^{3}$
	is assumed to be bounded and twice differentiable. \end{assum}

\subsection{Expanded Vehicle Nonlinear Dynamics}

By substituting \eqref{eq:Rmatrix} in \eqref{eq:eq1}, the expanded
state space equations of the UAV translation dynamics can be described
by: 
\begin{equation}
	\begin{cases}
		\dot{x}_{p} & =v_{1}\\
		\dot{v}_{1} & =-2m^{-1}F_{th}\left(q_{0}q_{2}+q_{1}q_{3}\right)\\
		\dot{y}_{p} & =v_{2}\\
		\dot{v}_{2} & =-2m^{-1}F_{th}\left(q_{2}q_{3}-q_{0}q_{1}\right)\\
		\dot{z}_{p} & =v_{3}\\
		\dot{v}_{3} & =-m^{-1}F_{th}\left(q_{0}^{2}-q_{1}^{2}-q_{2}^{2}+q_{3}^{2}\right)+g
	\end{cases}\label{eq:eq7}
\end{equation}
Expanding \eqref{eq:eq6} results in the following rotational dynamics:
\begin{equation}
	\begin{aligned}\dot{p}_{v} & =I_{xx}^{-1}(\left(I_{yy}-I_{zz}\right)q_{v}r_{v}+\mathcal{T}_{1})\\
		\dot{q}_{v} & =I_{yy}^{-1}(\left(I_{zz}-I_{xx}\right)r_{v}p_{v}+\mathcal{T}_{2})\\
		\dot{r}_{v} & =I_{zz}^{-1}(\left(I_{xx}-I_{yy}\right)p_{v}q_{v}+\mathcal{T}_{3})
	\end{aligned}
	\label{eq:eq8}
\end{equation}
Likewise, in the light of \eqref{eq:eq6}, the quaternion orientation
dynamics are given by: 
\begin{equation}
	\begin{aligned}\dot{q}_{0} & =-\frac{1}{2}\left(q_{1}p_{v}+q_{2}q_{v}+q_{3}r_{v}\right)\\
		\dot{q}_{1} & =\frac{1}{2}\left(q_{0}p_{v}-q_{3}q_{v}+q_{2}r_{v}\right)\\
		\dot{q}_{2} & =\frac{1}{2}\left(q_{3}p_{v}+q_{0}q_{v}-q_{1}r_{v}\right)\\
		\dot{q}_{3} & =\frac{1}{2}\left(-q_{2}p_{v}+q_{1}q_{v}+q_{0}r_{v}\right)
	\end{aligned}
	\label{eq:eqQ}
\end{equation}

\section{Problem Formulation and Controller Design\label{sec:Control}}

\begin{figure*}
	\centering{}\includegraphics[scale=0.4]{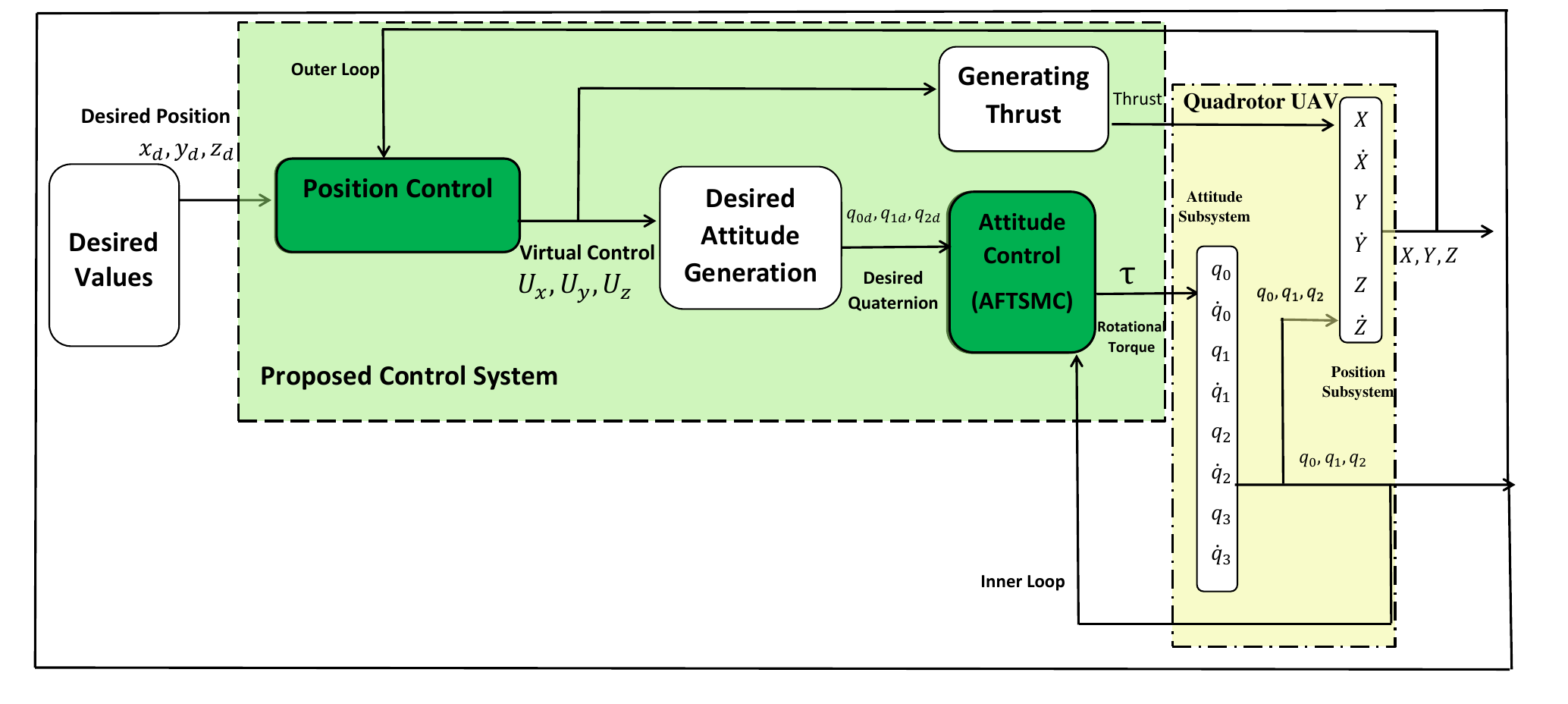}\caption{Illustrative diagram of the proposed control system for UAV}
	\label{fig:fig5} 
\end{figure*}

Solving singularity and chattering issues involves making autonomous
UAV safer, more efficient, and more integrated into diverse applications.
The main goal of this section is to design robust and singular-free
controllers for the UAV to follow predefined desired trajectories,
namely desired position $X_{d}=[x_{d},y_{d},z_{d}]^{\top}\in\mathbb{R}^{3}$
and desired attitude $Q_{d}=[q_{d0},q_{d1},q_{d2},q_{d3}]^{\top}\in\mathbb{S}^{3}$
guaranteeing UAV stability and asymptotic convergence in a finite
time. Tightly coupled cascaded control (to handle UAV underactuation)
is proposed where the outer control, translational part (position
and linear velocity control) is quaternion-based ABC while the inner
control, rotational part (orientation and angular velocity control)
is quaternion-based AFTSMC able to reduce chattering, guarantee asymptotic
stability, and ensure finite time convergence. The proposed controller
utilizes four control inputs to control seven states $(q_{0},q_{1},q_{2},q_{3},x_{p},y_{p},z_{p})$.
The proposed control system is shown in Figure \ref{fig:fig5}.

\subsection{Adaptive Backstepping for Position Control\label{subsec:pose}}

A quaternion-based ABC algorithm is designed for the UAV translation
to track predefined desired translation trajectories to resolve persistent
singularity issues in Euler-based controllers (e.g., \cite{labbadi2019robust}).
Backstepping is a superior technique for underactuated systems because
of its recursive approach for implementation \cite{zhao2020adaptive}.
The first step is to find the error between the UAV's true and desired
position. Based on \eqref{eq:eq7}, the error equations $e_{p}=[e_{xp},e_{yp},e_{zp}]^{\top}\in\mathbb{R}^{3}$
can be derived as follow: 
\begin{equation}
	\begin{aligned}e_{xp} & =x_{p}-x_{d}\\
		e_{yp} & =y_{p}-y_{d}\\
		e_{zp} & =z_{p}-z_{d}
	\end{aligned}
	\label{eq:eq9}
\end{equation}
such that the derivative of (\ref{eq:eq9}) is given by 
\begin{equation}
	\begin{aligned}\dot{e}_{xp} & =\dot{x}_{p}-\dot{x}_{d}=v_{1}-\dot{x}_{d}\\
		\dot{e}_{yp} & =\dot{y}_{p}-\dot{y}_{d}=v_{2}-\dot{y}_{d}\\
		\dot{e}_{zp} & =\dot{z}_{p}-\dot{z}_{d}=v_{3}-\dot{z}_{d}
	\end{aligned}
	\label{eq:eq10}
\end{equation}
Let us define the following Lyapunov function candidate and subsequently
formulate the controller that ensures system stability. Consider the
following Lyapunov function: 
\begin{equation}
	\begin{aligned}V_{1} & =\frac{1}{2}e_{xp}^{2}\\
		V_{3} & =\frac{1}{2}e_{yp}^{2}\\
		V_{5} & =\frac{1}{2}e_{zp}^{2}
	\end{aligned}
	\label{eq:eq11}
\end{equation}
In view of \eqref{eq:eq11} and with the substitution of \eqref{eq:eq10},
one obtains 
\begin{equation}
	\begin{aligned}\dot{V}_{1} & =e_{xp}\left(\dot{e}_{xp}\right)=e_{xp}\left(v_{1}-\dot{x}_{d}\right)\\
		\dot{V}_{3} & =e_{yp}\left(\dot{e}_{yp}\right)=e_{yp}\left(v_{2}-\dot{y}_{d}\right)\\
		\dot{V}_{5} & =e_{zp}\left(\dot{e}_{zp}\right)=e_{zp}\left(v_{3}-\dot{z}_{d}\right)
	\end{aligned}
	\label{eq:eq12}
\end{equation}
Therefore, $v_{1d}$, $v_{2d}$, and $v_{3d}$ should be selected
to stabilize \eqref{eq:eq12} as as follows: 
\begin{equation}
	\begin{aligned}v_{1d} & =-m_{xp}e_{xp}+\dot{x}_{d}\\
		v_{2d} & =-m_{yp}e_{yp}+\dot{y}_{d}\\
		v_{3d} & =-m_{zp}e_{zp}+\dot{z}_{d}
	\end{aligned}
	\label{eq:eq13}
\end{equation}
where $m_{xp}$, $m_{yp}$, and $m_{zp}$ are non-zero and non-negative
constants. Consider the translation dynamics \eqref{eq:eq7} and let
Assumption \ref{Assum:Assume1} and \ref{Assum:Assume2} hold true.
The position errors in \eqref{eq:eq9} converges to origin if the
desired $v_{1d}$, $v_{2d}$, and $v_{3d}$ are selected as \eqref{eq:eq13}.
It is obvious that $e_{xp}$, $e_{yp}$, and $e_{zp}$ are converging
to zero, if \eqref{eq:eq12} becomes negative definite. One finds
\begin{equation}
	\begin{aligned}\dot{V}_{1} & =e_{xp}\left(\dot{e}_{xp}\right)=e_{xp}\left(-m_{xp}e_{xp}+\dot{x}_{d}-\dot{x}_{d}\right)\\
		& =-m_{xp}e_{xp}^{2}<0\\
		\dot{V}_{3} & =e_{yp}\left(\dot{e}_{yp}\right)=e_{yp}\left(-m_{yp}e_{yp}+\dot{y}_{d}-\dot{y}_{d}\right)\\
		& =-m_{yp}e_{yp}^{2}<0\\
		\dot{V}_{5} & =e_{zp}\left(\dot{e}_{zp}\right)=e_{zp}\left(-m_{zp}e_{zp}+\dot{z}_{d}-\dot{z}_{d}\right)\\
		& =-m_{zp}e_{zp}^{2}<0
	\end{aligned}
	\label{eq:eq14}
\end{equation}
As a next step, we need to show that $v_{1}$, $v_{2}$, and $v_{3}$
are converging to $v_{1d}$, $v_{2d}$, and $v_{3d}$, respectively.
Let us define the following set of errors: 
\begin{equation}
	\begin{aligned}e_{v1} & =v_{1}-v_{1d}\\
		e_{v2} & =v_{2}-v_{2d}\\
		e_{v3} & =v_{3}-v_{3d}
	\end{aligned}
	\label{eq:eq15}
\end{equation}
Similar to the previous step, consider selecting the following Lyapunov
function candidate: 
\begin{equation}
	\begin{aligned}V_{2} & =V_{1}+\frac{1}{2}e_{v1}^{2}\\
		V_{4} & =V_{3}+\frac{1}{2}e_{v2}^{2}\\
		V_{6} & =V_{5}+\frac{1}{2}e_{v3}^{2}
	\end{aligned}
	\label{eq:eq16}
\end{equation}
In view of \eqref{eq:eq14} and \eqref{eq:eq16}, one obtains 
\begin{equation}
	\begin{aligned}\dot{V}_{2} & =\dot{V}_{1}+e_{v1}\dot{e}_{v1}=-m_{xp}e_{xp}^{2}+e_{v1}\left(\dot{v}_{1}-\dot{v}_{1d}\right)\\
		\dot{V}_{4} & =\dot{V}_{3}+e_{v2}\dot{e}_{v2}=-m_{yp}e_{yp}^{2}+e_{v2}\left(\dot{v}_{2}-\dot{v}_{2d}\right)\\
		\dot{V}_{6} & =\dot{V}_{5}+e_{v3}\dot{e}_{v3}=-m_{zp}e_{zp}^{2}+e_{v3}\left(\dot{v}_{3}-\dot{v}_{3d}\right)
	\end{aligned}
	\label{eq:eq17}
\end{equation}
where $m_{xp}$, $m_{yp}$, and $m_{zp}$ are positive constants to
be selected subsequently. One way to address the UAV underactuated
issue is by utilizing the backstepping approach to generate intermediary
control signals (virtual control). Now, $U_{x}$, $U_{y}$, and $U_{z}$
are introduced as virtual controls (inner control) to generate the
total thrust.  Substituting \eqref{eq:eq9}, \eqref{eq:eq10}, and
\eqref{eq:eq13} in \eqref{eq:eq16} and by using $U_{x}$, $U_{y}$,
and $U_{z}$ as virtual variables, the following equations are obtained
as: 
\begin{equation}
	\begin{aligned}\dot{V_{2}} & =-m_{xp}e_{xp}^{2}+e_{v1}\left(U_{x}+\left(m_{xp}\dot{e}_{xp}-\ddot{x}_{d}\right)\right)\\
		\dot{V_{4}} & =-m_{yp}e_{yp}^{2}+e_{v2}\left(U_{y}+\left(m_{yp}\dot{e}_{yp}-\ddot{y}_{d}\right)\right)\\
		\dot{V_{6}} & =-m_{zp}e_{zp}^{2}+e_{v3}\left(U_{z}+\left(m_{zp}\dot{e}_{zp}-\ddot{z}_{d}\right)\right)
	\end{aligned}
	\label{eq:eq18}
\end{equation}
By plugging $\dot{e}_{xp}$, $\dot{e}_{yp}$, and $\dot{e}_{zp}$
from \eqref{eq:eq10} and \eqref{eq:eq13}, the virtual control input
$U_{x}$, $U_{y}$, $U_{z}$ are designed as follows: 
\begin{equation}
	\begin{aligned}U_{x} & =m_{xp}^{2}e_{xp}-\hat{n}_{x2}e_{v1}+\ddot{x}_{d}\\
		U_{y} & =m_{yp}^{2}e_{yp}-\hat{n}_{y2}e_{v2}+\ddot{y}_{d}\\
		U_{z} & =m_{zp}^{2}e_{zp}-\hat{n}_{z2}e_{v3}+\ddot{z}_{d}
	\end{aligned}
	\label{eq:eq19}
\end{equation}
with $\hat{n}_{x2}$, $\hat{n}_{y2}$, and $\hat{n}_{z2}$ being
adaptive parameter estimation to automatically adjust position control
parameters in terms of stability and the speed of convergence and
their adaptation laws are given by: 
\begin{equation}
	\text{Update Law}\begin{cases}
		\dot{\hat{n}}_{x2} & =\eta_{1}e_{v1}^{2}\\
		\dot{\hat{n}}_{y2} & =\eta_{2}e_{v2}^{2}\\
		\dot{\hat{n}}_{z2} & =\eta_{3}e_{v3}^{2}
	\end{cases}\label{eq:ad1p}
\end{equation}
where $\eta_{1}$, $\eta_{2}$, and $\eta_{3}$ are positive constants. 
\begin{lem}
	\label{Lemm:lem1} \cite{labbadi2019robust} Based on Barbalet Lemma,
	if $f(s)$ is a uniformly bounded continuous function and $\lim_{s\to+\infty}\int_{0}^{\top}f(s)\,ds$
	exists, $f(s)$ converges to the origin asymptotically. 
\end{lem}
\begin{thm}
	\label{thm:Theorem1} The translation dynamics described in \eqref{eq:eq7}
	are asymptotically stable if the UAV thrust is designed as follows:
	\begin{equation}
		F_{th}=m\sqrt{U_{x}^{2}+U_{y}^{2}+\left(U_{z}+g\right)^{2}}\label{eq:eq23}
	\end{equation}
	where $U_{x}$, $U_{y}$, and $U_{z}$ are calculated as in \eqref{eq:eq19}
	with adaptation laws defined in \eqref{eq:ad1p}. 
\end{thm}
\textbf{Proof}: To prove that the subsystem is stable and to determine
$\hat{n}_{x2}$, $\hat{n}_{y2}$, and $\hat{n}_{z2}$, the Lyapunov
candidate of position system is selected. Lemma \ref{Lemm:lem1} \cite{labbadi2019robust}
is employed to ensure controller stability and error convergence.
By substituting for the intermediary control laws in \eqref{eq:eq19},
the stability of the translation system is guaranteed as follows:
\begin{equation}
	\begin{aligned}V_{tot}= & V_{2}+V_{4}+V_{6}\\
		= & \frac{1}{2}e_{xp}^{2}+\frac{1}{2}e_{yp}^{2}+\frac{1}{2}e_{zp}^{2}+\frac{1}{2}e_{v1}^{2}+\frac{1}{2}e_{v2}^{2}+\frac{1}{2}e_{v3}^{2}\\
		& +\frac{1}{2}\tilde{n}_{x2}^{2}+\frac{1}{2}\tilde{n}_{y2}^{2}+\frac{1}{2}\tilde{n}_{z2}^{2}
	\end{aligned}
	\label{eq:eq20}
\end{equation}
where $\tilde{n}_{x2}=\hat{n}_{x2}-n_{x2}$, $\tilde{n}_{y2}=\hat{n}_{y2}-n_{y2}$,
and $\tilde{n}_{z2}=\hat{n}_{z2}-n_{z2}$. Based on \eqref{eq:eq18},
the derivative of \eqref{eq:eq20} is computed as follows: 
\begin{equation}
	\begin{aligned}\dot{V}_{tot} & =\dot{V}_{2}+\dot{V}_{4}+\dot{V}_{6}+\tilde{n}_{x2}\dot{\tilde{n}}_{x2}+\tilde{n}_{y2}\dot{\tilde{n}}_{y2}+\tilde{n}_{z2}\dot{\tilde{n}}_{z2}\end{aligned}
	\label{eq:d20}
\end{equation}
By inserting \eqref{eq:eq19} in \eqref{eq:eq18} and \eqref{eq:d20},
the above equation is simplified as: 
\begin{equation}
	\begin{aligned}\dot{V}_{tot} & =-m_{xp}e_{xp}^{2}-m_{yp}e_{yp}^{2}-m_{zp}e_{zp}^{2}-\hat{n}_{x2}e_{v1}^{2}-\hat{n}_{y2}e_{v2}^{2}\\
		& -\hat{n}_{z2}e_{v3}^{2}+\tilde{n}_{x2}\dot{\hat{n}}_{x2}+\tilde{n}_{y2}\dot{\hat{n}}_{y2}+\tilde{n}_{z2}\dot{\hat{n}}_{z2}
	\end{aligned}
	\label{eq:eq20d}
\end{equation}
Hence, by inserting \eqref{eq:ad1p} in the derivative of \eqref{eq:eq20d},
the stability can be guaranteed such that: 
\begin{equation}
	\begin{aligned}\dot{V}_{tot} & =-m_{xp}e_{xp}^{2}-m_{yp}e_{yp}^{2}-m_{zp}e_{zp}^{2}-(\tilde{n}_{x2}+n_{x2})e_{v1}^{2}\\
		& -(\tilde{n}_{y2}+n_{y2})e_{v2}^{2}-(\tilde{n}_{z2}+n_{z2})e_{v3}^{2}+\tilde{n}_{x2}e_{v1}^{2}+\tilde{n}_{y2}e_{v2}^{2}\\
		& +\tilde{n}_{z2}e_{v3}^{2}=-m_{xp}e_{xp}^{2}-m_{yp}e_{yp}^{2}-m_{zp}e_{yp}^{2}-n_{x2}e_{v1}^{2}\\
		& -n_{y2}e_{v2}^{2}-n_{z2}e_{v3}^{2}\leq0
	\end{aligned}
	\label{eq:eq21}
\end{equation}
As \eqref{eq:eq21} becomes negative definite for all $V_{tot}>0$
and $\dot{V}_{tot}=0$ only at $V_{tot}=0$, it is ensured that the
errors converge to the origin and the system is asymptotically stable.

Let us calculate the total thrust ($F_{th}$) necessary to control
the UAV translation dynamics and generate the desired attitude trajectories.
Based on \eqref{eq:eq7} and \eqref{eq:eq19}, the virtual controls
$U_{x}$, $U_{y}$, and $U_{z}$ can be re-expressed as : 
\begin{equation}
	\begin{aligned}U_{x} & =-2m^{-1}F_{th}\left(q_{0}q_{2}+q_{1}q_{3}\right)\\
		U_{y} & =-2m^{-1}F_{th}\left(q_{2}q_{3}-q_{0}q_{1}\right)\\
		U_{z} & =-m^{-1}F_{th}\left(q_{0}^{2}-q_{1}^{2}-q_{2}^{2}+q_{3}^{2}\right)+g
	\end{aligned}
	\label{eq:eq22}
\end{equation}
Solving for \eqref{eq:eq22} one obtains the actual thrust $F_{th}$
as follows: 
\begin{equation}
	F_{th}=m\sqrt{U_{x}^{2}+U_{y}^{2}+\left(U_{z}+g\right)^{2}}\label{eq:eq233}
\end{equation}
Based on \eqref{eq:eq233}, it becomes obvious that the actual total
thrust will not be subject to singularity at any time instant. In
view of \eqref{eq:eq22} and \eqref{eq:eq23}, the desired quaternion
$q_{0d}$, $q_{1d}$, $q_{2d}$ as follows:

\begin{equation}
	\begin{aligned}q_{0d} & =\sqrt{\frac{m\,\left(g+U_{z}\right)}{2\,F_{th}}+\frac{1}{2}}\\
		q_{1d} & =-\frac{m\,U_{y}}{2\,F_{th}\,q_{0d}}\\
		q_{2d} & =\frac{m\,U_{x}}{2\,F_{th}\,q_{0d}}
	\end{aligned}
	\label{eq:eq24}
\end{equation}
visit \cite{hashim2023observer}. Note that it is assumed $q_{3d}=0$.
As a summary, in this section, quaternion-based ABC (outer control)
is designed for the UAV translation dynamics with an objective of
generating the total UAV thrust and desired quaternion components
for attitude control. In the next section, a quaternion-based AFTSMC
(inner control) will be developed for the UAV attitude control problem.

\subsection{Fast Terminal Sliding Mode Control Attitude Control\label{subsec:atitude}}

In comparison with the conventional SMC, AFTSMC brings more advantages
in terms of performance. AFTSMC significantly decreases chattering,
which is a crucial factor for smooth trajectory tracking and actuator
safety. Moreover, AFTSMC can address singularity issues of conventional
SMC guaranteeing finite time convergence \cite{mobayen2015adaptive}.
Consider the following attitude errors \cite{hashim2019special}:
\begin{equation}
	\begin{aligned}e_{0} & =q_{0}q_{0d}+\boldsymbol{q}_{d}\boldsymbol{q}^{\top}\\
		e & =q_{0d}\boldsymbol{q}-q_{0}\boldsymbol{q}_{d}+[\boldsymbol{q}]_{\times}\boldsymbol{q}_{d}\\
		R_{e} & =RR_{d}^{\top}=(e_{0}^{2}-e^{\top}e)\mathbf{I}_{3}+2ee^{\top}+2e_{0}[e]_{\times}
	\end{aligned}
	\label{eq:eq25}
\end{equation}
where $e_{0}\in\mathbb{R}$ and $e\in\mathbb{R}^{3}$ denoted quaternion
error between UAV true and desired trajectories, $R\in SO(3)$ and
$R_{d}$ refers to true and desired rotation matrix of the UAV, respectively,
and $R_{e}\in SO(3)$ denotes error in orientation. The goal is to
drive $R\rightarrow R_{d}$ or in other words $R_{e}\rightarrow\mathbf{I}_{3}$.
The equivalent unit-quaternion mapping is to drive $e_{0}\rightarrow1$
and $e\rightarrow0_{3\times1}$. Let us define the error in angular
velocity is defined by: 
\begin{equation}
	\Omega_{e}=\Omega_{B}-R_{e}\Omega_{d}\label{eq:eq27}
\end{equation}
where the desired angular velocity $\Omega_{d}$ can be found using
the following equation \cite{hashim2019special}: 
\begin{equation}
	\boldsymbol{\dot{Q}_{d}}=(q_{0d}\mathbf{I}_{3}+[\boldsymbol{q}_{d}]_{\times})\Omega_{d}\label{eq:eq28}
\end{equation}
with $\boldsymbol{q}_{d}=[q_{1d},q_{2d},q_{3d}]^{\top}$. Based on
\eqref{eq:eq25},  \eqref{eq:eq27} and \eqref{eq:eq6}, the attitude
error dynamic is obtained as \cite{hashim2023observer}: 
\begin{equation}
	\text{Error dynamic}\begin{cases}
		\dot{e}_{0} & =\frac{1}{2}e^{\top}\Omega_{e}\\
		\dot{e} & =\frac{1}{2}\left(e_{0}\Omega_{e}+[e]_{\times}\Omega_{e}\right)\\
		J\dot{\Omega}_{e} & =[J\Omega_{B}]_{\times}\Omega_{B}+\mathcal{T}+\\
		& \hspace{0.3cm}J\left[\Omega_{e}\right]_{\times}R_{e}\Omega_{d}-JR_{e}\dot{\Omega}_{d}
	\end{cases}\label{eq:eq29}
\end{equation}
Let us move to the Quaternion-based AFTSMC and introduce the following
sliding surface \eqref{eq:eq29}: 
\begin{equation}
	s=\dot{e}+\gamma_{1}e+\gamma_{2}e^{n/l}+\gamma_{3}\Omega_{e}\label{eq:eq30}
\end{equation}
where $\gamma_{1}$ and $\gamma_{2}$ are positive constants and $\gamma_{3}=1$.
$n$ and $l$ are positive constants and $0<\frac{n}{l}<1$. The derivative
of \eqref{eq:eq30} is calculated as: 
\begin{equation}
	\dot{s}=\ddot{e}+\gamma_{1}\dot{e}+\frac{n}{l}\gamma_{2}e^{(n/l-1)}\dot{e}+\dot{\Omega}_{e}\label{eq:eq31}
\end{equation}
Based on \eqref{eq:eq31}, it is obvious that when $e$ becomes zero,
$e^{(\frac{n}{l}-1)}$ could converge to infinity. In such case, the
system could be subject to singularity and thereby become unstable.
In view of \cite{labbadi2019robust,hua2019fractional}, to resolve
this issue, the sliding surface described in \eqref{eq:eq30} can
be modified by switching threshold value to overcome the singularity
issue as follows: 
\begin{equation}
	s=\dot{e}+\gamma_{1}e+\gamma_{2}\Delta(e)+\gamma_{3}\Omega_{e}\label{eq:eq32}
\end{equation}
where $\Delta(e)$ is defined as: 
\begin{equation}
	\Delta\left(e\right)=\begin{cases}
		e^{\frac{n}{l}}, & \text{ if }s=0\text{ or }s\neq0,\left|e\right|>\epsilon\\
		e, & \text{ if }s\neq0,\left|e\right|\leqslant\epsilon
	\end{cases}\label{eq:eq33}
\end{equation}
with $\epsilon$ being a small positive value. Based on \eqref{eq:eq32}
and \eqref{eq:eq33}, let us propose the attitude controller as follows:
\begin{equation}
	\begin{aligned}\mathcal{T}= & -([J\Omega_{B}]_{\times}\Omega_{B}+J\left[\Omega_{e}\right]_{\times}R_{e}\Omega_{d}-JR_{e}\dot{\Omega}_{d})\\
		& -(\frac{1}{2}e_{0}+\frac{1}{2}[e]_{\times}\Omega_{e}+\mathcal{I})^{-1}(\frac{1}{2}\dot{e_{0}}\Omega_{e}+\frac{1}{2}[\dot{e}]_{\times}\Omega_{e}\\
		& +\gamma_{1}\dot{e}+\gamma_{2}(n/l)e^{n/l-1}\dot{e}+\mu_{1}s+\boldsymbol{\hat{K}}sign(s))
	\end{aligned}
	\label{eq:eq34}
\end{equation}
where $\mu_{1}$ is a positive constant and the adaptive law for $\boldsymbol{\hat{K}}$=$[\hat{k}_{1},\hat{k}_{2},\hat{k}_{3}]^{\top}$
is as follows: 
\begin{equation}
	\text{Update Law}\begin{cases}
		\dot{\hat{k}}_{1} & =\lambda||s||\\
		\dot{\hat{k}}_{2} & =\lambda||s||\\
		\dot{\hat{k}}_{3} & =\lambda||s||
	\end{cases}\label{eq:ad1}
\end{equation}
where $\lambda>0$. 
\begin{thm}
	\label{thm:Theorem2} The attitude system \eqref{eq:eq6} becomes
	asymptotically stable and the attitude error $Q_{e}=[e_{0},e^{\top}]^{\top}\rightarrow[1,0,0,0]^{\top}$
	and $\Omega_{e}\rightarrow0$ when the proposed input controller is
	defined as \eqref{eq:eq34} with the adaptation laws in \eqref{eq:ad1}. 
\end{thm}
\textbf{Proof}: The Lyapunov approach and Lemma \ref{Lemm:lem1} are
used to prove Theorem \ref{thm:Theorem2}. Let us introduce the following
Lyapunov function candidate: 
\begin{equation}
	V_{s}=\frac{1}{2}s^{\top}s+\frac{1}{2\lambda}\tilde{\boldsymbol{K}}^{\top}\tilde{\boldsymbol{K}}\label{eq:eq35}
\end{equation}
where $\tilde{\boldsymbol{K}}=\hat{\boldsymbol{K}}-\boldsymbol{K}$.
The derivative of \eqref{eq:eq35} is obtained by: 
\begin{equation}
	\begin{aligned}\dot{V}_{s}=s\left(\ddot{e}+\gamma_{1}\dot{e}+\gamma_{2}\frac{n}{l}e^{n/l-1}\dot{e}+\dot{\Omega}_{e}\right)+\frac{1}{\lambda}\tilde{\boldsymbol{K}}^{\top}\dot{\tilde{\boldsymbol{K}}}\end{aligned}
	\label{eq:eq36}
\end{equation}
Also, the $\ddot{e}$ can be derived from \eqref{eq:eq29} as: 
\begin{equation}
	\begin{aligned} & \ddot{e}=\frac{1}{2}(\dot{e}_{0}\Omega_{e}+e_{0}\dot{\Omega}_{e}+[\dot{e}]_{\times}\Omega_{e}+[e]_{\times}\dot{\Omega}_{e})\end{aligned}
	\label{eq:eq37}
\end{equation}
Using \eqref{eq:eqOne}, \eqref{eq:eq37} and \eqref{eq:eq29}, the
above equation can be rewritten as follows: 
\begin{equation}
	\begin{aligned} & \dot{V}_{s}=s(\frac{1}{2}((\dot{e}_{0}\mathbf{I}_{3}+[\dot{e}]_{\times})\Omega_{e}+(e_{0}\mathbf{I}_{3}+[e]_{\times})\dot{\Omega}_{e})+\\
		& \frac{\gamma_{1}}{2}\left(e_{0}\Omega_{e}+[e]_{\times}\Omega_{e}\right)+\gamma_{2}\frac{n}{l}e^{n/l-1}\dot{e}+J^{-1}([J\Omega_{B}]_{\times}\Omega_{B}+\\
		& \mathcal{T}+J\left[\Omega_{e}\right]_{\times}R_{e}\Omega_{d}-JR_{e}\dot{\Omega}_{d}))+\frac{1}{\lambda}\tilde{\boldsymbol{K}}^{\top}\dot{\hat{\boldsymbol{K}}}
	\end{aligned}
	\label{eq:eq38}
\end{equation}
By substituting \eqref{eq:eq34} in \eqref{eq:eq38}, the derivative
of \eqref{eq:eq35} becomes: 
\begin{equation}
	\begin{aligned} & \dot{V}_{s}=-\mu_{1}||s||^{2}-s\hat{\boldsymbol{K}}sign(s)+\frac{1}{\lambda}\tilde{\boldsymbol{K}}^{\top}\dot{\hat{\boldsymbol{K}}}\end{aligned}
	\label{eq39}
\end{equation}
Based on \eqref{eq:ad1}, the above equation can be rewritten as:
\begin{equation}
	\begin{aligned}\dot{V}_{s} & =-\mu_{1}||s||^{2}-s(\tilde{\boldsymbol{K}}+\boldsymbol{K})sign(s)+\frac{1}{\lambda}\tilde{\boldsymbol{K}}^{\top}\lambda||s||\\
		& \leq-\mu_{1}||s||^{2}-\boldsymbol{K}||s||\leq0
	\end{aligned}
	\label{eq:ad2}
\end{equation}
Therefore, the attitude dynamics in \eqref{eq:eq8} become asymptotically
stable using the proposed controller in \eqref{eq:eq34} completing
the proof.

The proposed quaternion-based AFTSMC guarantees asymptotic convergence
of the attitude dynamics. However, finite time convergence is not
guaranteed. This issue will be addressed in the subsequent subsection.

\subsection{Attitude Control and Finite Time Convergence}

In this subsection, the finite time $t_{r}$ of the AFTSMC is computed
for the proposed controller \eqref{eq:eq34}. It is assumed that the
time interval between the initial error values $[e(0)\neq0,\Omega_{e}(0)\neq0]\neq0$
to reach $e=0$ is $t_{r}$. 
\begin{lem}
	\label{lemm:lem3} The state errors reach the AFTSMC surface in \eqref{eq:eq32}
	using the controller \eqref{eq:eq34} and adaptive update law \eqref{eq:ad1}
	in a finite time $t_{r}$. 
\end{lem}
\textbf{Proof}: To obtain finite time convergence, consider the following
real value function: 
\begin{equation}
	V_{tr}=\frac{1}{2}s^{2}\label{eq:t1}
\end{equation}
The derivative of \eqref{eq:t1} is obtained as: 
\begin{equation}
	\begin{aligned}\dot{V}_{tr}=s\left(\ddot{e}+\gamma_{1}\dot{e}+\gamma_{2}\frac{n}{l}e^{n/l-1}\dot{e}+\dot{\Omega}_{e}\right)\end{aligned}
	\label{eq:t2}
\end{equation}
Using \eqref{eq:eq34} and \eqref{eq:eqOne}, the above expression
\eqref{eq:t2} can be rewritten as follows: 
\begin{equation}
	\begin{aligned} & \dot{V}_{tr}=s(\frac{1}{2}((\dot{e}_{0}\mathbf{I}_{3}+[\dot{e}]_{\times})\Omega_{e}+(e_{0}\mathbf{I}_{3}+[e]_{\times})\dot{\Omega}_{e})+\\
		& \gamma_{1}(\frac{1}{2}\left(e_{0}\Omega_{e}+[e]_{\times}\Omega_{e}\right))+\gamma_{2}\frac{n}{l}e^{n/l-1}\dot{e}+J^{-1}([J\Omega_{B}]_{\times}\Omega_{B}+\\
		& \mathcal{T}+J\left[\Omega_{e}\right]_{\times}R_{e}\Omega_{d}-JR_{e}\dot{\Omega}_{d}))
	\end{aligned}
	\label{eq:t3}
\end{equation}
Substituting for \eqref{eq:eq34} in \eqref{eq:t3}, one obtains:
\begin{equation}
	\begin{aligned} & \dot{V}_{tr}\leqslant-\frac{1}{2}\mu_{1}s^{2}-\mu_{2}|s|\leqslant0\end{aligned}
	\label{eq:t4}
\end{equation}
such that 
\begin{equation}
	\dot{V}_{tr}=\frac{dV_{tr}}{dt}\leq-\mu_{1}V_{tr}-\mu_{2}\sqrt{2}V_{tr}^{1/2}\label{eq:t5}
\end{equation}
Let us introduce the variable $\varrho=\mu_{2}\sqrt{2}$. One finds:
\begin{equation}
	dt\leq\frac{-dV_{tr}}{\mu_{1}V_{tr}+\varrho V_{tr}^{1/2}}\label{eq:t6}
\end{equation}
such that 
\begin{equation}
	dt\leq\frac{-V_{tr}^{-1/2}dV_{tr}}{\mu_{1}V_{tr}^{1/2}+\varrho}=-2\frac{dV_{tr}^{1/2}}{\mu_{1}V_{tr}^{1/2}+\varrho}\label{eq:t7}
\end{equation}
Integrating both sides of \eqref{eq:t7}, the finite time can be derived
as: 
\begin{equation}
	\begin{aligned}\int_{0}^{t_{r}}dt & \leq\int_{V_{tr}(0)}^{V_{tr}\left(t_{r}\right)}\frac{-2dV_{tr}^{1/2}}{\mu_{1}V_{tr}^{1/2}+\varrho}\\
		& =\left[\frac{-2}{\mu_{1}}\ln\left(\mu_{1}V_{tr}^{1/2}+\varrho\right)\right]_{V_{tr}(0)}^{V_{tr}\left(t_{r}\right)}
	\end{aligned}
	\label{eq:t8}
\end{equation}
Finally, based on \eqref{eq:t8}, $t_{r}$ is calculated as: 
\begin{equation}
	t_{r}\leq\frac{2}{\mu_{1}}\ln\left(\frac{\mu_{1}V_{tr}(0)^{1/2}+\varrho}{\varrho}\right)\label{eq:t9}
\end{equation}
According to \eqref{eq:t9}, it is concluded that sliding surface
$s$ in \eqref{eq:eq35} converges to zero, and states reach the desired
values in finite time. Also, as long as $s=0$, then error output
and states converge to zero in finite time as well.

Finally, both pose and attitude quaternion-based are obtained in \ref{subsec:pose}
and \ref{subsec:atitude} sections, respectively. Stabilizing the
UAV system and enhancing following the desired position and attitude
trajectories are achieved by designing proposed controllers. The flight
control diagram, which includes backstepping and AFTSMC, is shown
in Figure \ref{fig:fig5}. In the following section, the algorithm
steps to design controllers are presented to facilitate controller
implementation for the UAV.

\section{Implementation Steps\label{sec:sum}}

The implementation process of the proposed UAV controller is presented
step by step in this section as follows:

\textbf{Step 1.} Define the initial values $X_{0}=[x_{p}(0),y_{p}(0),z_{p}(0)]$,
$Q_{0}=[q_{0}(0),q_{1}(0),q_{2}(0),q_{3}(0)]^{\top}$, and $\Omega_{B_{0}}=[p_{v}(0),q_{v}(0),r_{v}(0)]$.
Calculate for $R$ and $R_{d}$ as follows: 
\begin{equation}
	\begin{aligned}R & =\left(q_{0}^{2}-\boldsymbol{q}^{\top}\boldsymbol{q}\right)I_{3}+2\boldsymbol{q}\boldsymbol{q}^{\top}+2q_{0}\mathbf{[q]}_{\times}\\
		R_{d} & =\left(q_{d0}^{2}-\boldsymbol{q}_{d}^{\top}\boldsymbol{q}_{d}\right)I_{3}+2\boldsymbol{q}_{d}\boldsymbol{q}_{d}^{\top}+2q_{d0}[\mathbf{q}_{d}]_{\times}
	\end{aligned}
\end{equation}

\textbf{Step 2.} Set UAV's desired position trajectory: $X_{d}=[x_{d},y_{d},z_{d}]^{\top}$

\textbf{Step 3.} Calculate position errors using \eqref{eq:eq9}:
\begin{align*}
	e_{xp} & =x_{p}-x_{d}\\
	e_{yp} & =y_{p}-y_{d}\\
	e_{zp} & =z_{p}-z_{d}
\end{align*}
and follow \eqref{eq:eq13}, \eqref{eq:eq14} to design virtual $v_{1_{d}}$,
$v_{2_{d}}$, and $v_{3_{d}}$: 
\begin{align*}
	v_{1d} & =-m_{xp}e_{xp}+\dot{x}_{d}\\
	v_{2d} & =-m_{yp}e_{yp}+\dot{y}_{d}\\
	v_{3d} & =-m_{zp}e_{zp}+\dot{z}_{d}
\end{align*}
and find: 
\begin{align*}
	e_{v1} & =v_{1}-v_{1d}\\
	e_{v2} & =v_{2}-v_{2d}\\
	e_{v3} & =v_{3}-v_{3d}
\end{align*}

\textbf{Step 4.} Design and implement the ABC control for translation
dynamics with the update laws in \eqref{eq:eq19} and \eqref{eq:ad1p}:
\begin{align*}
	U_{x} & =-m_{xp}^{2}e_{xp}-\hat{n}_{x2}e_{v1}+\ddot{x}_{d}\\
	U_{y} & =-m_{yp}^{2}e_{yp}-\hat{n}_{y2}e_{v2}+\ddot{y}_{d}\\
	U_{z} & =-m_{zp}^{2}e_{zp}-\hat{n}_{z2}e_{v3}+\ddot{z}_{d}
\end{align*}
where the control parameter adaptation to compensate for uncertainties
are given by: 
\[
\text{Adaptive update}\begin{cases}
	\dot{\hat{n}}_{x2} & =\eta_{1}e_{v1}^{2}\\
	\dot{\hat{n}}_{y2} & =\eta_{2}e_{v2}^{2}\\
	\dot{\hat{n}}_{z2} & =\eta_{3}e_{v3}^{2}
\end{cases}
\]

\textbf{Step 5.} Find the actual thrust using the previous step as
follows: 
\[
F_{th}=m\sqrt{U_{x}^{2}+U_{y}^{2}+\left(U_{z}+g\right)^{2}}
\]

\textbf{Step 6.} Generate the desired attitude and angular velocity
using \eqref{eq:eq27} and \eqref{eq:eq31} as follows: 
\begin{align*}
	q_{0d} & =\sqrt{\frac{m\,\left(g+U_{z}\right)}{2\,F_{th}}+\frac{1}{2}}\\
	q_{1d} & =-\frac{m\,U_{y}}{2\,F_{th}\,q_{0d}}\\
	q_{2d} & =\frac{m\,U_{x}}{2\,F_{th}\,q_{0d}}\\
	q_{3d} & =0
\end{align*}

\textbf{Step 7.} Calculate attitude errors using \eqref{eq:eq28}
to \eqref{eq:eq30}: 
\begin{align*}
	e_{0} & =q_{0}q_{0d}+\boldsymbol{q}_{d}\boldsymbol{q}^{\top}\\
	e & =q_{0d}\boldsymbol{q}-q_{0}\boldsymbol{q}_{d}+[\boldsymbol{q}]_{\times}\boldsymbol{Q_{d}}\\
	R_{e} & =RR_{d}^{\top}
\end{align*}

\textbf{Step 8.} Build the AFTSMC sliding surface using \eqref{eq:eq36}
and \eqref{eq:eq37} as: 
\[
s=\dot{e}+\gamma_{1}e+\gamma_{2}\Delta(e)+\gamma_{3}\Omega_{e}
\]
\begin{figure*}
	\centering{}\includegraphics[scale=0.3]{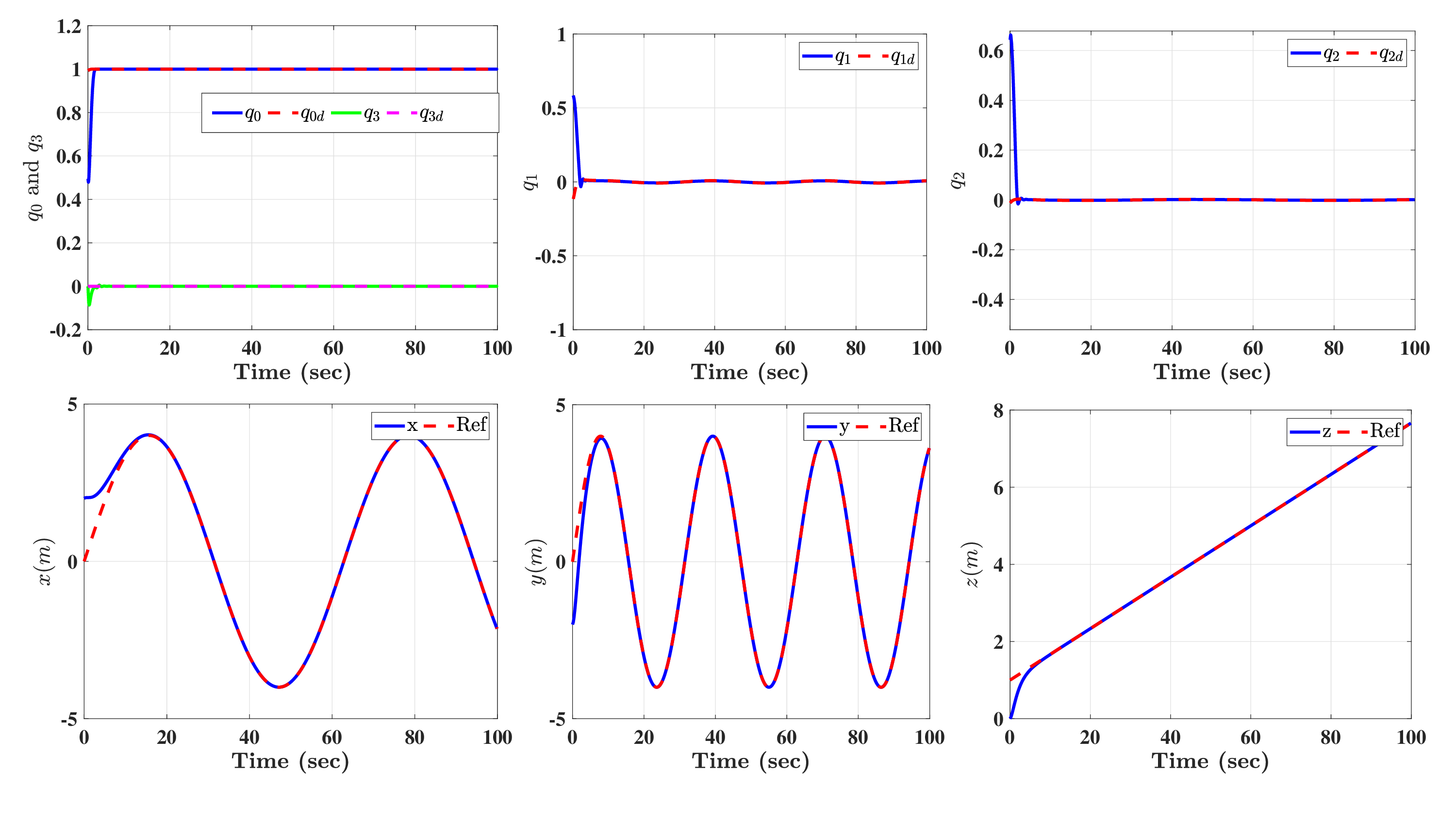} \caption{UAV Attitude and Position Trajectory Tracking: True value represented
		in the blue solid-line, while red dash-lines plotted references}
	\label{fig:fig1} 
\end{figure*}
\textbf{Step 9.} Implement AFTSMC attitude controller using \eqref{eq:eq37}
and the attitude adaptive law \eqref{eq:ad1}. 
\begin{align*}
	\mathcal{T} & =-([J\Omega_{B}]_{\times}\Omega_{B}+J\left[\Omega_{e}\right]_{\times}R_{e}\Omega_{d}-JR_{e}\dot{\Omega}_{d})\\
	& -(\frac{1}{2}e_{0}+\frac{1}{2}[e]_{\times}\Omega_{e}+\mathcal{I})^{-1}(\frac{1}{2}\dot{e_{0}}\Omega_{e}+\frac{1}{2}[\dot{e}]_{\times}\Omega_{e}\\
	& +\gamma_{1}\dot{e}+\gamma_{2}(n/l)e^{n/l-1}\dot{e}+\mu_{1}s+\boldsymbol{\hat{K}}sign(s))
\end{align*}
with Adaptive update to minimize chattering: 
\[
\text{Adaptive update}\begin{cases}
	\dot{\hat{k}}_{1} & =\lambda||s||\\
	\dot{\hat{k}}_{2} & =\lambda||s||\\
	\dot{\hat{k}}_{3} & =\lambda||s||
\end{cases}
\]

\textbf{Step 10.} Once the rotational torque ($\mathcal{T}=[\tau_{1},\tau_{2},\tau_{3}]^{\top}$)
and thrust ($F_{th}$) are obtained, the required drone rotor speeds
are calculated by \cite{hashim2023exponentially}: 
\[
\left[\begin{array}{c}
	\omega_{1}^{2}\\
	\omega_{2}^{2}\\
	\omega_{3}^{2}\\
	\omega_{4}^{2}
\end{array}\right]=\left[\begin{array}{cccc}
	c_{d} & -c_{d} & c_{d} & -c_{d}\\
	-l_{d}b_{d} & 0 & l_{d}b_{d} & 0\\
	0 & -l_{d}b_{d} & 0 & l_{d}b_{d}\\
	b_{d} & b_{d} & b_{d} & b_{d}
\end{array}\right]^{-1}\left[\begin{array}{c}
	\tau_{1}\\
	\tau_{2}\\
	\tau_{3}\\
	F_{th}
\end{array}\right]
\]
where $c_{d}$ refers to drag coefficient, $l_{d}$ denotes the distance
between the quadrotor center of mass, $b_{d}$ refers to thrust factor,
and $\omega_{1}$, $\omega_{2}$, $\omega_{3}$, and $\omega_{4}$
refer to rotors speed.

\textbf{Step 11.} To address the chattering problem, one can replace
$sign(\cdot)$ function (discontinuous part) with the $\tanh(\cdot)$
function. Repeat Steps 3 to 10.

\section{Simulation Results\label{sec:Result}}

In this section, the simulation is conducted to demonstrate the effectiveness
and robustness of the proposed quaternion-based controllers. We study
the proposed quaternion-based ABC and AFTSMC performance for the translational
and rotational dynamic of the underactuated quadrotor UAV. Firstly,
the proposed control system is performed to verify its performance
in handling unknown time-varying uncertainties. Subsequently, we present
comparative simulation outcomes to demonstrate the superiority, stability,
and singularity-free of the developed quaternion-based controllers
compared to the Euler-based established method in the literature.
Table. \ref{tab:Table3} provides physical system and control design
parameters.



\subsection{Simulation Scenario}

In this section, the simulation is conducted to verify the robustness
of the proposed quaternion-based controllers by applying unknown time-varying
uncertainties of parameters such as $0.025sin(0.03t+0.3)$, $0.03sin(0.02t+0.25)$,
$0.04sin(0.04t+0.35)$, and $0.035sin(0.015t+0.4)$. We study the
proposed quaternion-based ABC and AFTSMC performance for the translational
and rotational dynamic of the underactuated quadrotor UAV. Firstly,
the proposed control system is performed to verify its performance
in handling unknown time-varying uncertainties. Subsequently, we present
comparative simulation outcomes to demonstrate the superiority, stability,
and singularity-free of the developed quaternion-based controllers
compared to the Euler-based established method in the literature.
\begin{figure*}
	\centering{}\includegraphics[scale=0.3]{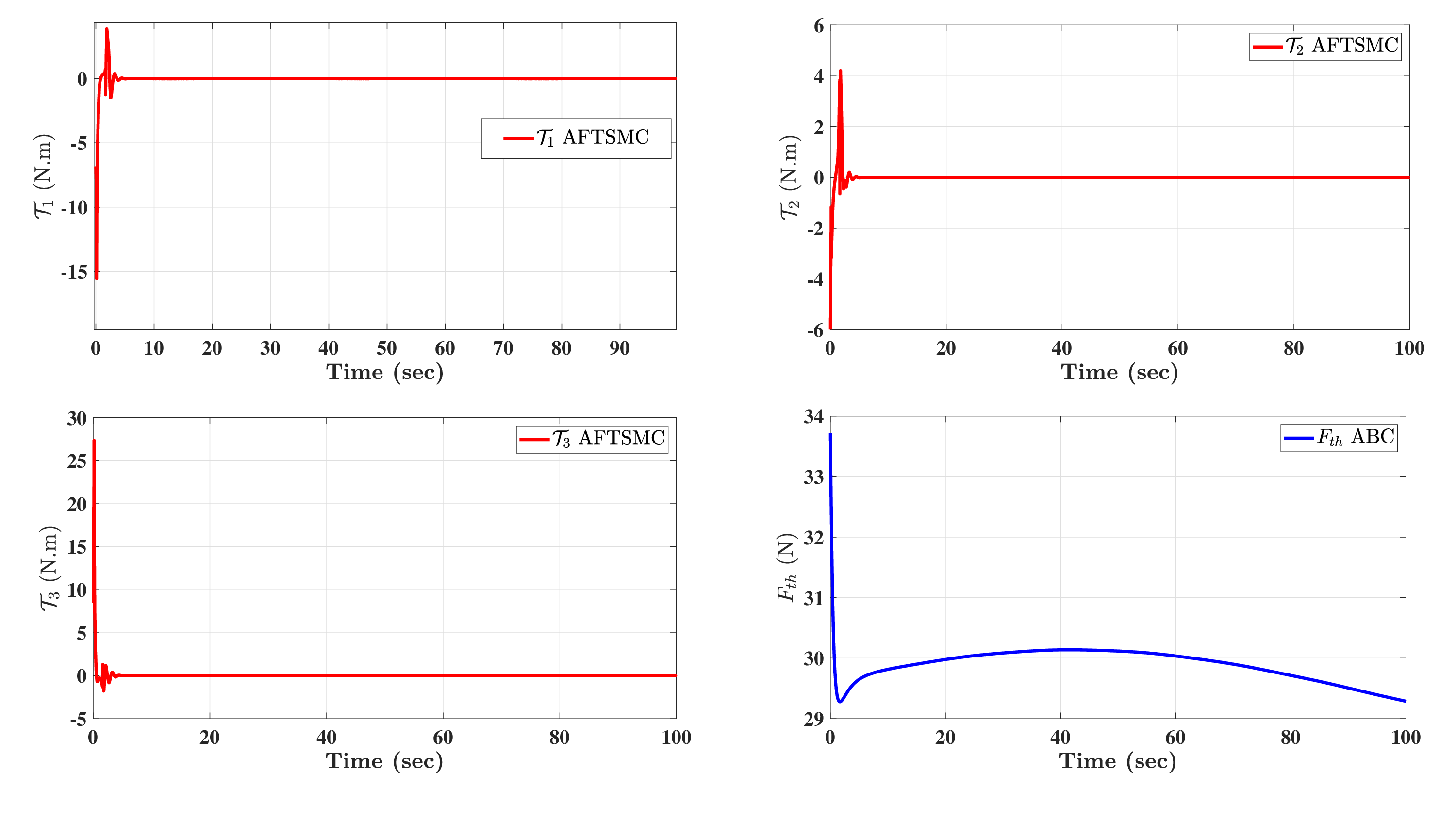} \caption{UAV Rotational Control (plotted in red) and Total
		Thrust Input (represented in blue).}
	\label{fig:fig2} 
\end{figure*}

\begin{figure*}
	\centering{}\includegraphics[scale=0.3]{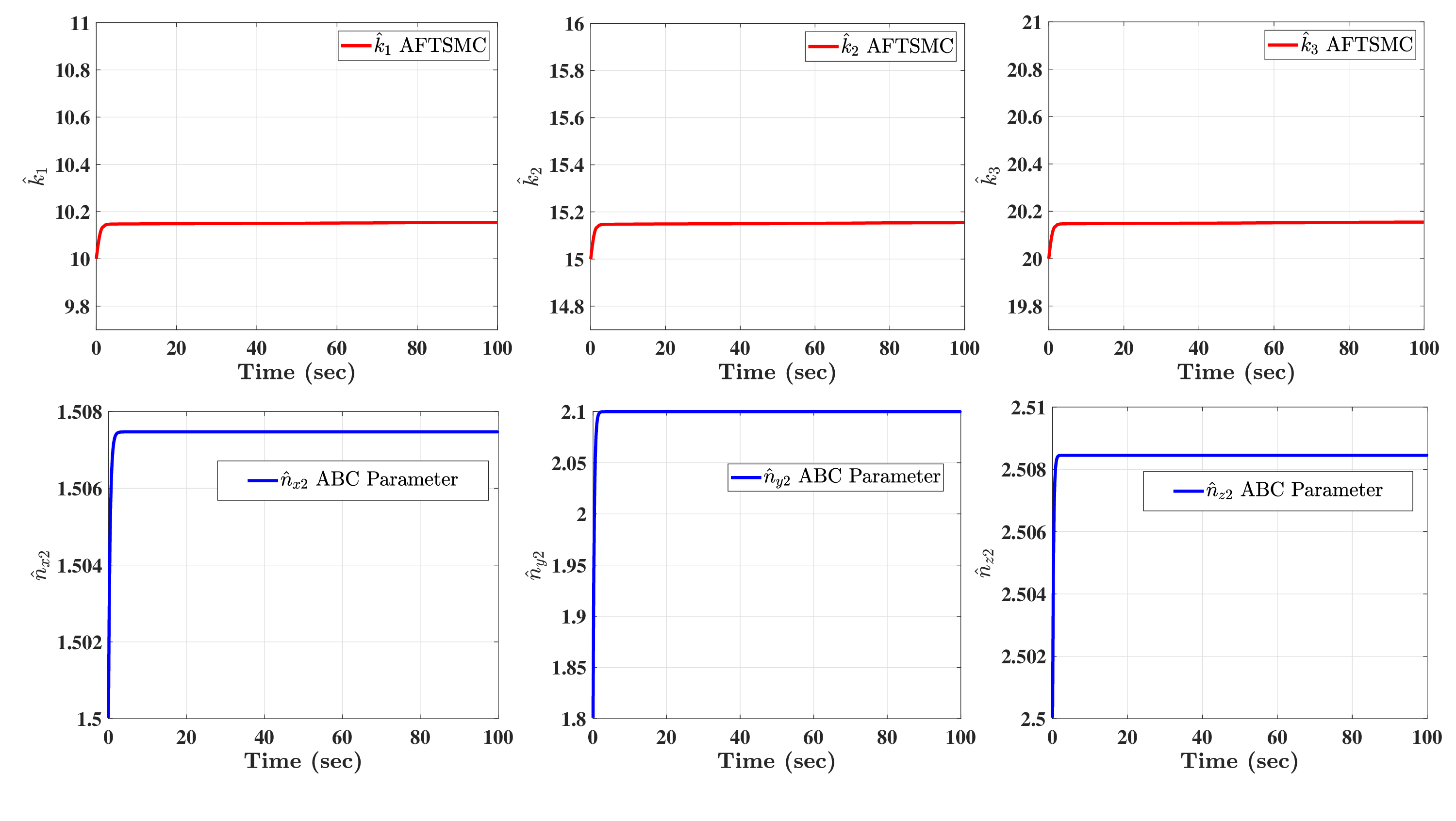} \caption{Bounded and smooth trajectory of the adaptive parameters
		of controllers.}
	\label{fig:fig3} 
\end{figure*}
\begin{table}[t]
	\centering{}\caption{\label{tab:Table3}Physical system and control design parameters.}
	
	\begin{tabular}{ll|ll}
		\hline 
		Parameters  & Value  & Parameters  & Value\tabularnewline
		\hline 
		$m_{xp}$  & 0.3  & m  & 3kg\tabularnewline
		$m_{yp}$  & 0.5  & $I_{xx}$  & $1.5kg.m^{2}$\tabularnewline
		$m_{zp}$  & 0.6  & $I_{yy}$  & $1.5kg.m^{2}$\tabularnewline
		$\gamma_{1}$  & 10  & $I_{zz}$  & $3kg.m^{2}$\tabularnewline
		$\gamma_{2}$  & 30  & g  & $9.8m/s^{2}$\tabularnewline
		$\gamma_{3}$  & 1  & $x_{d}$  & $4cos(0.1t)$\tabularnewline
		$\eta_{1}=\eta_{2}=\eta_{3}$  & 0.01  & $y_{d}$  & $4cos(0.2t)$\tabularnewline
		$n$  & 3  & $z_{d}$  & $(1/15)t+1$\tabularnewline
		$l$  & 5  & $\lambda$  & 30\tabularnewline
		\hline 
	\end{tabular}
\end{table}

\begin{figure*}
	\centering{}\includegraphics[scale=0.3]{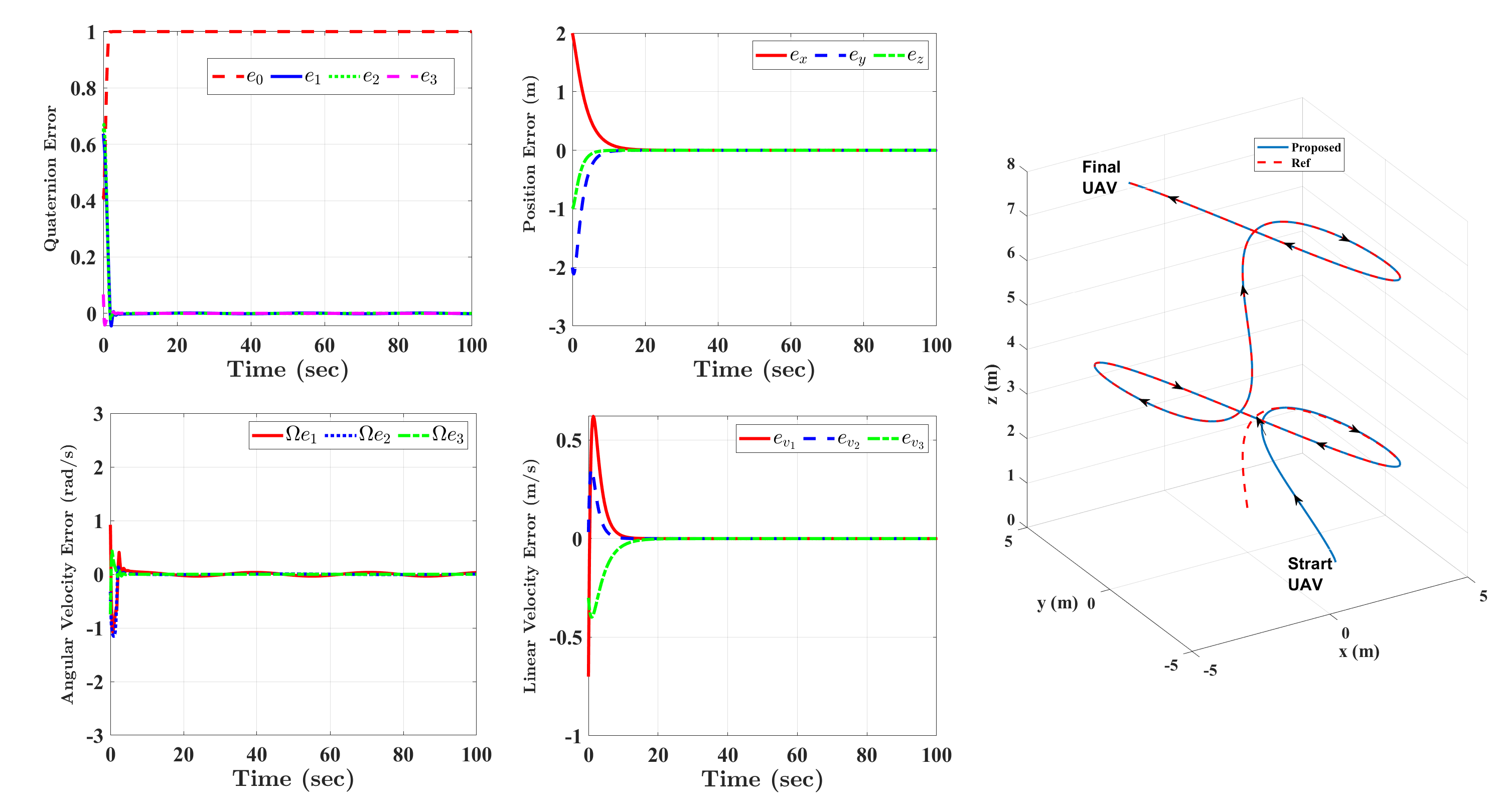}\caption{UAV flight trajectories and Errors of the proposed
			quaternion-based AFTSMC.}
	\label{fig:fig4} 
\end{figure*}

The performance of the proposed controllers is shown in Figure \ref{fig:fig1}-\ref{fig:fig4}.
In Figure \ref{fig:fig1}, the attitude and position trajectories
follow desired values; meanwhile, attitude desired values are generated
by total thrust to address underactuated complexity of the UAV. Figure
\ref{fig:fig2} presents the attitude control $\mathcal{T}$ and total
thrust $F_{th}$, respectively. According to Figure \ref{fig:fig2},
the controller is robust and smooth. Also, the quaternion-based AFTSMC
is robust to uncertainties without singularity and chattering issues.
Figure \ref{fig:fig3} shows adaptive control parameters from \eqref{eq:ad1}
and \eqref{eq:ad1p}. Figure \ref{fig:fig3} presents bounded adaptive
estimates. The quaternion error, position error, angular velocity
error, linear velocity error, and flight trajectory of UAV are shown
in Figure \ref{fig:fig4}. The error components converge successfully
and smoothly to the origin.

\begin{figure*}
	\centering{}\includegraphics[scale=0.3]{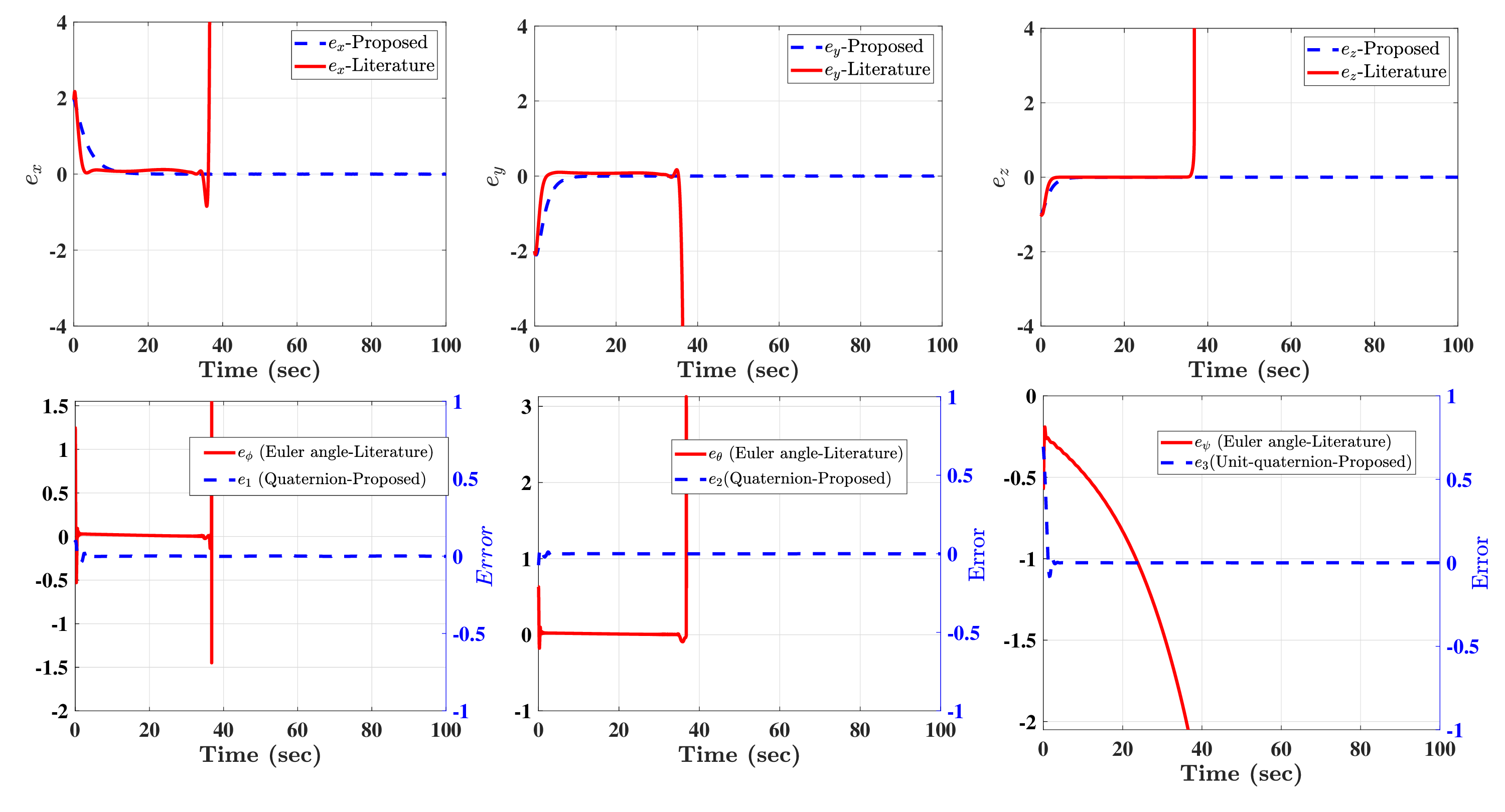}\caption{Position and attitude error comparison between literature \cite{labbadi2019robust} and proposed work: Euler angles-based from
			literature \cite{labbadi2019robust} plotted in red solid line vs
			the proposed quaternion-based ABC with AFTSMC approach plotted in
			blue dash-line.}
	\label{fig:fig7} 
\end{figure*}

The simulation results demonstrated in this paper validate the effectiveness
of the proposed controllers, particularly in handling unknown time-varying
uncertainties. The UAV performed excellent tracking capabilities and
stability during the mission, as depicted in Figure \ref{fig:fig1}.
Theorem \ref{thm:Theorem1} and Lemma \ref{Lemm:lem1} guarantee the
asymptotic stability of the UAV's translation dynamics, which confirms
the simulation results shown in Figure \ref{fig:fig1}. The figure
depicts that the UAV's position is stable and smoothly tracks the
desired trajectories with strong accuracy while significant initialization
is applied. Moreover, Theorem \ref{thm:Theorem2} validates the convergence
of the UAV's attitude states, as also shown by Figure \ref{fig:fig1}.
The quaternion orientations effectively follow the desired values
generated by the thrust to address the underactuated complexity, and
the attitude tracking under control law in equation \eqref{eq:ad1}
achieves asymptotic stability.

Results in Figure \ref{fig:fig2} depict that the proposed quaternion-based
AFTSMC is highly effective in mitigating chattering issues, and its
performance is outstanding in bringing states to the desired values
in finite time. Additionally, the total thrust successfully adjusts
and adapts its performance to produce accurate and sufficient desired
attitude orientations for rotational control, while the underactuated
UAV works under parameter uncertainties. To handle the uncertainties,
the adaptive features of the proposed controller update control parameters
as shown in Figure \ref{fig:fig3}. It confirms that the updated control
parameters remain bounded and realistic. Also, $\hat{K}$ in \eqref{eq:eq34}
is to reduce chattering in the attitude controller, while $\hat{n}_{x2}$,
$\hat{n}_{y2}$, and $\hat{n}_{z2}$ have a critical role in the speed
of convergence and stability. The 6 DOF UAV flight trajectory exhibits
not only smoothness but also accuracy. As illustrated in Figure \ref{fig:fig4},
the developed controllers successfully achieve the defined goals,
including trajectory tracking and minimizing chattering issues. Notably,
the position and linear velocity errors converge to zero regardless
of significant initial conditions. Additionally, the Unit-quaternion
orientations converge the desired values of {[}$1,0,0,0${]}; meanwhile,
angular velocity errors attain zero. The outstanding outcomes confirm
the efficacy of the sliding surface defined in \eqref{eq:eq32}, as
it brings the system states to the desired values in finite time.

\subsection{Comparison}

A comprehensive comparison is performed to demonstrate the state-of-the-art
proposed quaternion-based controllers' performance over the Euler
angle-based counterparts (e.g.,\cite{labbadi2019robust}). The main
shortcoming of Euler angle-based controllers stems from kinematic
and model representation singularities, which often result in issues
such as system instability and control failure. To highlight the differences,
Figure \ref{fig:fig7} depicts the assessment of the attitude and
position errors of both approaches. The results provide valuable insights
into the advantages of using the proposed quaternion-based controller
in UAV flight control, validating robustness and effectiveness in
overcoming the deficiencies associated with Euler angle-based approaches.

Results demonstrate the success of the proposed quaternion-based controllers
in accurately tracking attitude and position trajectories. In contrast,
the Euler angle-based controllers exhibit limitations, failing to
capture orientation in certain configurations. The validation further
establishes that employing the Euler angle-based control system may
lead to instability, and in some cases, it becomes kinematic singular,
as depicted in Figure \ref{fig:fig7}. Conversely, the innovative
quaternion-based control system effectively addresses singularities,
providing a robust and globally singularity-free model for quadrotors.
Indeed, the validated superiority performance of the quaternion-based
approach emphasizes its remarkable ability to address the limitations
of Euler angle-based methods for controlling UAV flight. The successful
mitigation of singularities and improved model representation make
the proposed quaternion-based control system a viable solution for
enhancing UAV performance and safety during flight missions.

\section{conclusion\label{sec:co}}

This paper proposed a novel quaternion-based attitude and position
control for the underactuated UAV. The newly proposed quaternion-based
Adaptive Backstepping Control (ABC) addressed the complexity of controlling
the underactuated Unmanned Aerial Vehicle (UAV) system. The developed
quaternion-based Adaptive Fast Terminal Sliding Mode Control (AFTSMC)
was conducted to mitigate the chattering issue and guarantee finite
time convergence of orientations and angular velocity. Precise tracking
and asymptotic stability of translational and rotational dynamics
were guaranteed by utilizing the Lyapunov theorem and Barbalet Lemma.
The novel adaptive feature of controllers successfully handled the
persistent unknown-time varying parameter uncertainties by adjusting
control parameters. In fact, designing controllers based on unit-quaternion
addressed the kinematic and model singularity of other established
approaches (e.g., Euler angles). Simulation results validated the
efficacy of the proposed quaternion-based controllers in terms of
stability, singularity, and chattering issues. Moreover, the results
showed the outstanding performance of ABC and AFTSMC in accurate tracking
in the presence of parameter uncertainties.

\section{Disclosure Statement}

No potential conflict of interest was reported by the author(s).

\section{Data Availability Statement}

The authors confirm that the data supporting the findings of this
study are available within the article and its supplementary materials.
The codes are are available from the corresponding author upon reasonable
request.

\bibliographystyle{IEEEtran}
\bibliography{RefVTOL}

\end{document}